\documentclass[lettersize,journal]{IEEEtran}
\usepackage{amsmath,amsfonts}
\usepackage{algorithmic}
\usepackage{algorithm}
\usepackage{array}
\usepackage[caption=false,font=normalsize,labelfont=sf,textfont=sf]{subfig}
\usepackage{textcomp}
\usepackage{stfloats}
\usepackage{url}
\usepackage{verbatim}
\usepackage{graphicx}
\usepackage{cite}
\usepackage{booktabs}
\usepackage{subfig}
\usepackage{amsmath}
\usepackage{amssymb}
\usepackage{multirow}
\usepackage{slashbox}
\usepackage{bm}
\usepackage{paralist}
\usepackage{color, xcolor, colortbl}

\def\BibTeX{{\rm B\kern-.05em{\sc i\kern-.025em b}\kern-.08em
    T\kern-.1667em\lower.7ex\hbox{E}\kern-.125emX}}
\usepackage{balance}

\usepackage[square,numbers,sort&compress]{natbib}
\usepackage[font=small]{caption}

\def\widththird{0.30\linewidth}
\def\hspacefigure{\hspace{0.02in}}

\newcommand{\CUT}[1]{}
\newcommand{\rwh}[1]{\textcolor{black}{#1}}

\newcommand{\new}[1]{\textcolor{black}{#1}}

\begin{document}


\title{Joint Counting, Detection and Re-Identification for
Multi-Object Tracking}

\author{Weihong Ren,
        Denglu Wu,
        Hui Cao, 
		Xi’ai Chen,
		Zhi Han,
		and Honghai Liu, \IEEEmembership{Fellow,~IEEE}
\thanks{Weihong Ren and Honghai Liu are with the
School of Mechanical Engineering and Automation, State Key Lab of Robotics and Systems, Harbin Institute of Technology, Shenzhen 518055,
China. (e-mail: renweihong@hit.edu.cn, honghai.liu@hit.edu.cn) 


Denglu Wu is with the Shenzhen Duorou Intelligent System Co., Ltd., Shenzhen 518055, China (e-mail: wdlcas@gmail.com).

Hui Cao is with the School of Electrical Engineering, Xi’an Jiaotong University, Xi’an 710049, China (e-mail: huicao@mail.xjtu.edu.cn).

Xi'ai Chen and Zhi Han are with State Key Laboratory of Robotics, Shenyang Institute of Automation, Chinese Academy of Science, Beijing 110169, China. (e-mail: chenxiai@sia.cn, hanzhi@sia.cn)}
		}


\markboth{Journal of \LaTeX\ Class Files,~Vol.~14, No.~8, August~2021}%
{Shell \MakeLowercase{\textit{et al.}}: A Sample Article Using IEEEtran.cls for IEEE Journals}


\maketitle

\begin{abstract}
The recent trend in 2D multiple object tracking~(MOT) is jointly solving detection and tracking, where object detection and appearance feature (or motion) are learned simultaneously. Despite competitive
performance, in crowded scenes, joint detection and tracking usually fail to find accurate object associations due to missed or false detections. In this paper, we jointly model counting, detection and re-identification in an end-to-end framework, named CountingMOT, tailored for crowded scenes. 
\new{By imposing mutual object-count constraints between detection and counting, the CountingMOT tries to find a balance between object detection and crowd density map estimation, which can help it to recover missed detections or reject false detections.}
Our approach is an attempt to bridge the gap of object detection, counting, and re-Identification.
This is in contrast to prior MOT methods that either ignore the crowd density and thus are prone to failure in crowded scenes, or depend on local correlations to build a graphical relationship for matching targets. The proposed MOT tracker can perform online and real-time tracking, and achieves the state-of-the-art results on public benchmarks \emph{MOT16}~(MOTA of 79.7\%), \emph{MOT17}~(MOTA of 81.3\%) and \emph{MOT20}~(MOTA of 78.9\%). \textcolor{blue}{Source code is available at https://github.com/weihong9/CountingMOT.}
\end{abstract}


\section{Introduction}\label{sec:introduction}
\IEEEPARstart{M}{}ultiple object tracking (MOT) is an essential task for computer vision, and has been applied to many applications, such as industrial surveillance and autonomous driving. Tracking-by-detection has been a dominant paradigm for MOT for a long time, where object trajectories are obtained by associating object detections over a video through appearance features. Object detection~\cite{duan2019centernet,liu2020deep,zhao2019object} and \new{re-identification (reID)~\cite{hou2020iaunet,zhou2019person,ye2021deep,gu2022motion}} have achieved significant progress in recent years, but \emph{tracking-by-detection} paradigm can hardly perform real-time tracking due to the separately compute-intensive models. Thus, the recent trend in MOT is jointly solving detection and tracking~(JDT), where object detections and appearance features (or motions) are learned simultaneously. Despite many years of effort, JDT usually fails to find accurate object associations in crowd scenes due to missed or false detections. 

The crowd density map was proposed for object counting~\cite{lempitsky2010learning}, where the sum over a region in the map corresponds to the object count in that region. It predicts the object count without explicitly detecting objects, and thus is a reliable and informative clue in crowd scenes. Using multi-task learning technique, we propose a novel MOT approach, to jointly perform counting, detection and reID, especially for crowd scenes. We use object count from crowd density map to constrain the number of objects from detection task, which can help detection task to find missed detections. In return, we also use object detections to refine the quality of crowd density map, improving its localization ability. By imposing the mutual object-count constraints between detection and counting, the proposed model can recover missed detections or reject false detections. 
Besides, our approach can perform online and real-time multi-object tracking.

\begin{figure}[t]
\centering
\includegraphics[width=0.95\linewidth]{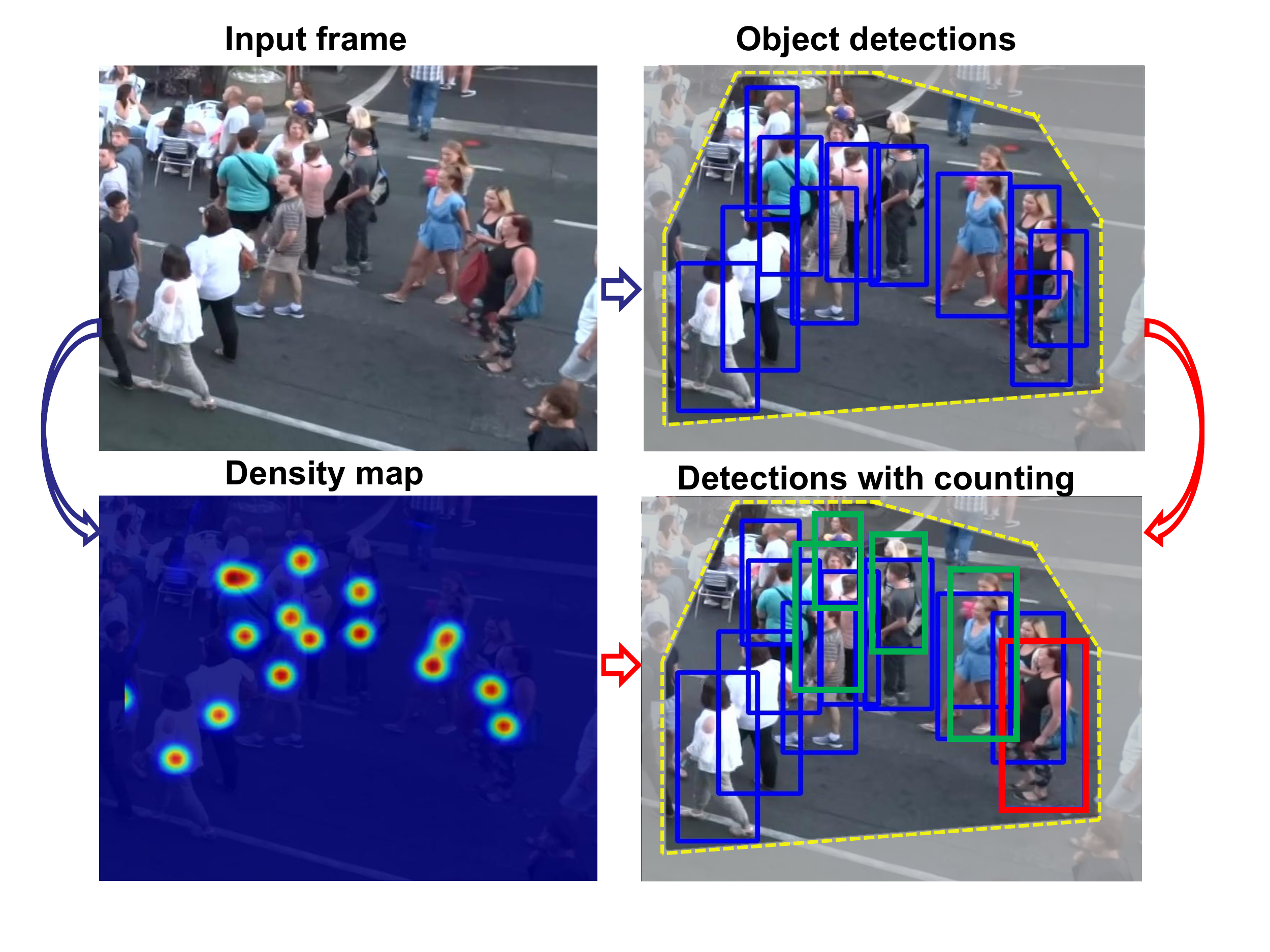}
\caption{Object detections with counting constraint in a crowd scene of~\emph{MOT20}. For clear visualization, we only show part of the object detections in the scene. Object detections in the top-right are generated by the state-of-the-art method~FairMOT~\cite{zhang2021fairmot}, which jointly produces object detections and reID features. However, FairMOT fails to locate occluded people in extremely crowd regions. By incorporating crowd density map (bottom-left) as a counting constraint, our proposed CountingMOT finds missed object detections (green boxes in the bottom-right) and also can eliminate false detections (the red box in the bottom-right).
}\label{fig:CountingMOT}
\vspace{-0.10in}
\end{figure}


In Fig.~\ref{fig:CountingMOT}, we show a qualitative comparison between the state-of-the-art MOT approach FairMOT~\cite{zhang2021fairmot}
and our proposed CountingMOT. FairMOT is a typical JDT method which produces object detections and reID features simultaneously. However, in extremely crowded regions, it fails to locate occluded people and also generates some false detections (see top-right), resulting in failures of data association. By incorporating the crowd density map, an informative clue in crowd scenarios (see bottom-left), into the object detection task, the proposed CountingMOT can find the missed object detections. Also, our method can correct false detections using the object count constraint from density map. E.g., in the right corner of the scene, there are four people together, and FairMOT misses an object and produces two inaccurate boxes in one person. Using the counting constraint from density map (bottom-left), our CountingMOT can recover the missed object, and the false detections also are eliminated.


Previous works~\cite{dehghan2017binary, ren2020tracking} have explored crowd density maps for multi-object tracking. They first predict the crowd density maps, and then rely on either a joint objective function or a discriminative model with initialized object detections for tracking.    
The approach of~\cite{dehghan2017binary} uses appearance, motion and contextual information to solve multi-object tracking in high-density crowd scenes, but it needs initial object detections in the first frame and then predicts the potential location of each object in the following frames by a quadratic objective function. Also, the method doesn't have an effective strategy to detect new objects, and thus usually works well for structure scenes. The recent method~\cite{ren2020tracking} proposes a \emph{tracking-by-counting} paradigm, which jointly models detection, and counting of multiple people as a network flow problem. It can achieve the global optimal detections and trajectories over a pre-given sequence of crowd density maps, but building a dense graph over the crowd density maps is time-consuming. Also, the tracking performance depends heavily on the crowd density maps. In contrast to~\cite{dehghan2017binary,ren2020tracking}, our proposed approach is an end-to-end deep learning model, which simultaneously produces object detections, crowd density map, and reID features, and can perform online and 
real-time tracking. The counting task and the object detection task can boost each other by using the same object-count constraint within a given area, which makes our approach robust to crowd scenes.


Therefore, our contribution is an end-to-end MOT framework that jointly solves counting, object detection and re-identification simultaneously. By imposing mutual object-count constraints between detection and counting, the two tasks can be optimized and enhanced at the same time, making the proposed approach robust to crowd scenes. \new{The proposed CountingMOT tries to find a balance between object detection and crowd density map estimation, which can help it to recover missed detections or reject false detections, and thus the MOT performance can be improved.}
Our approach is an attempt to bridge the gap of object detection, counting, and re-Identification.
The experimental results demonstrate the superiority of our approach against the state-of-the-arts on public benchmarks.

{The remainder of this paper is organized as follows. Section \ref{text:related} reviews the previous MOT works, including \emph{tracking-by-detection} and \emph{joint detection-and-tracking}. Section \ref{text:tbc} introduces our proposed CountingMOT, and the experiments on public benchmark datasets are conducted in Section \ref{text:experiments}. Finally, Section \ref{text:conclusion} concludes this work.}

\section{Related work}
\label{text:related}
In this section, we briefly review the related works on \emph{tracking-by-detection} MOT approaches, \emph{joint detection-and-tracking} MOT
approaches and crowd counting approaches. Comprehensive reviews on MOT can be found in~\cite{ciaparrone2020deep,dendorfer2021motchallenge}.

\subsection{MOT approaches using tracking-by-detection}
Tracking-by-detection is a standard paradigm for MOT, where the problem is split into two stages: object detection and data association. Here, we generally divide \emph{tracking-by-detection} approaches into two categories, i.e., batch basis for offline scenarios
~\cite{yu2017adaptive,schulter2017deep,son2017multi,braso2020learning} 
and frame-by-frame basis for online applications~\cite{chu2017online,zhu2018online,guo2021online}.

Early tracking-by-detection approaches regard data association as a global optimization problem \new{using batch input}, and various formulations are proposed, such as continuous energy optimization~\cite{milan2015multi}, min-cost network flow~\cite{butt2013multi}, 
and Conditional Random Field
(CRF)~\cite{yang2011learning}. \new{However, the above traditional methods need to manually build a flow graph with hand-crafted features or costs, which is cumbersome for tracking. 
Recently, some approaches~formulate the network flow of MOT into a fully differentiable neural network, which can adaptively learn features or costs for data association. E.g., \cite{schulter2017deep} presents a deep network flow that expresses the optimum of network flow as a differentiable function of pairwise association costs, and it can perform data association in an end-to-end fashion.
\cite{braso2020learning}~constructs a flow graph of MOT using CNNs, where the nodes represent object detections and the edges indicate the associations across different frames. \cite{xiang2020end} also formulates the assignment costs as unary and pairwise potentials, and then uses a recurrent neural network to gradually refine tracklet association.
Deep flow approaches can improve the tracking performance using the powerful neural networks, but the inability to real-time tracking shifts the further research~\cite{gao2021crf}.}
Using batch detections as input, the MOT methods usually transform data association \new{as an offline energy optimization problem.} Though they can use long-term trajectory to recover missed or occluded detections, the batch processing strategy can hardly be applied to realtime applications. 

\new{To perform online and realtime tracking}, \cite{bewley2016simple} is a primary attempt towards online tracking where only detections from the previous and the current frames are presented. The MOT is solved by an assignment problem with the Hungarian algorithm where the assignment cost matrix is formulated by intersection-over-union (IOU). This method runs very fast, and it also indicates that tracking performance is highly dependent on detection results. To reduce Identity Switches (IDS), \cite{wojke2017simple} further extends \cite{bewley2016simple} by incorporating appearance information,  making it a strong MOT tracker at high frame rates. \cite{yu2016poi} also proves that high-performance detection and deep appearance feature can lead to the state-of-the-art multi-object tracking results. 
\new{Recent trend in MOT is to leverage the powerful representational ability of deep learning~\cite{
sun2019deep,dai2021learning} to perform online data association.} 
Using recurrent neural networks (RNNs), \cite{milan2017online}~formulates data association and trajectory estimation into a
neural network without tuning tedious hyper-parameters. However, this method can't run in real time. 
\cite{sun2019deep} proposed a deep affinity network to infer object affinities across different frames, and it is a realtime tracker with high tracking performance. 
\new{Single object tracking (SOT) has achieved great advances, and MOT approaches can benefit from the development of SOT~\cite{yuan2020self,zheng2021improving}.} \cite{zheng2021improving}~extends the detection network by adding a
SOT branch for tracking objects, making the MOT task have the powerful discrimination ability of SOT. 
Based on the siamese tracker, 
\cite{shuai2021siammot}~proposes a region-based MOT network to simultaneously \new{detect and associate} object instances. 
Using graph convolutional network (GCN), \cite{dai2021learning}~models MOT as a proposal generation and trajectory inference problem, \new{but} it still faces the problem of occlusion in crowd scenes. The above methods can improve tracking performance using the advances of deep learning, \new{but they are subjected to wrong detections in crowd scenes, which is a key factor for MOT~\cite{bewley2016simple,yu2016poi,shuai2021siammot}}.

To handle missed detections or occlusions, 
\cite{xiao2015collaborative}~adopts a hierarchical
tracking system using different priorities to resolve long-term occlusion.
 \cite{fu2019multi}~integrates full body and body parts to address ambiguous identity associations for people tracking. 
\cite{wang2021dynamic}~introduces a dynamic tracking system, which maintains tracking results for each frame by combining global and local search. 
The above methods can handle crowd scenes well, but they still follow the tracking-by-detection paradigm, which limits the tracking efficiency in real applications.

\subsection{MOT approaches with joint detection-and-tracking}

The \emph{tracking-by-detection} MOT approaches \new{usually focus on real-time data association, but 
pay little attention to the detection step and thus are not real-time MOT systems~\cite{zhang2021fairmot}}. Recently, many efforts aim to solve object detection, feature embedding or motion prediction simultaneously~\cite{lu2020retinatrack,zhou2020tracking,bergmann2019tracking,pang2020tubetk,
wang2021joint,pang2021quasi,wu2021track,wang2021multiple}.
\cite{wang2020towards} proposes a pioneering work that allows object detection and appearance embedding to be learned in a unified model, but the embedding is generated from a positive anchor which may shift to neighbouring objects in crowd scenes. Further, \cite{lu2020retinatrack}~modifies a one-stage detector to capture the instance-level embedding, but it still suffers the identity ambiguity problem when two anchors are centered at the same grid.
To address this ambiguity, \cite{zhang2021fairmot}~learns appearance embedding in an anchor-free approach, i.e., extracting reID features at the
object center. \cite{bergmann2019tracking} converts 
an object detector into a MOT tracker, by exploiting the bounding box regression branch to predict the new object positions in the next frame, but its performance is limited by the object detector. Also, \cite{zhou2020tracking} extends a classic detector CenterNet~\cite{duan2019centernet} to a MOT tracker which directly uses an additional branch to predict object motions, but it lacks the modelling of object appearance. Built on~\cite{zhou2020tracking}, \cite{tokmakov2021learning}~takes pairs of frames as input, allowing it to recover occluded object detections or trajectories using historical information, but it is not applicable to extremely crowd scenes. 
To realize one-step tracking, \cite{pang2020tubetk}~directly predicts bounding tubes using spatial-temporal information in overlapped video clips. Using graph neural networks, \cite{wang2021joint}~models both detection and data association in a relational graph, and it can improve object detection results using temporal relations. \cite{wu2021track}~proposes two modules to jointly learn appearance embedding and object motions, it proves that tracking clue enhances object detection and in return benefits object tracking. To distinguish similar objects, \cite{wang2021multiple}~presents a correlation tracking model that  exploits the temporal context, which can make the trajectory temporarily consistent. The above \emph{joint detection-and-tracking} approaches either utilize the temporal information to recover trajectories, or focus on learning discriminative embeddings. They may ignore the most important thing in MOT, i.e., object detection which is the dominant factor for tacking.

\begin{figure*}[!htbp]
\centering
\includegraphics[width=0.85\linewidth]{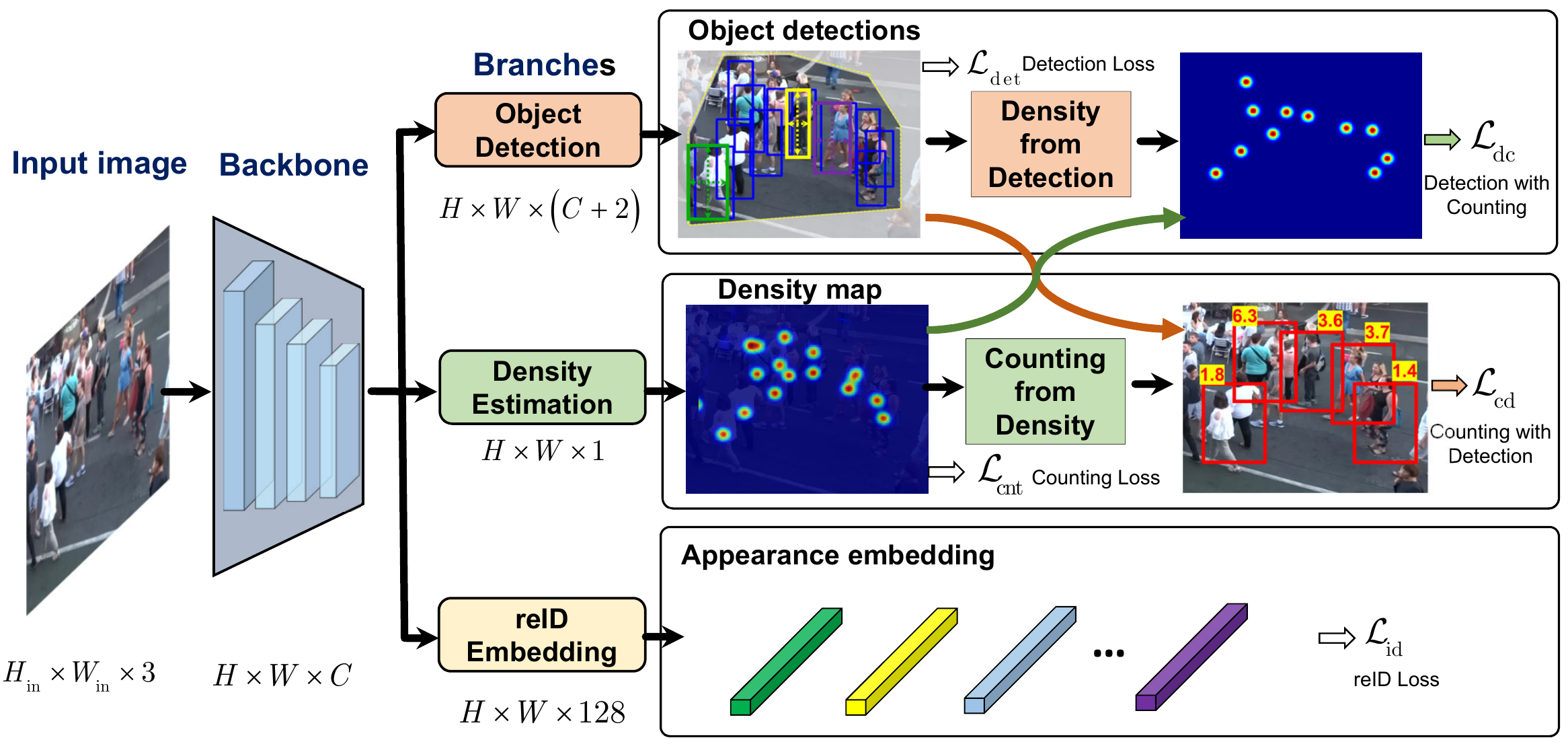}
\caption{The proposed CountingMOT model for joint Counting, Detection, and re-Identification. 
The input image is first fed to the backbone for multi-level feature extraction. 
Then, we add three homogeneous branches for simultaneously performing detection, counting and reID, respectively. Also, we create mutual constraints between detection and counting to improve detections in crowd scenes. The reID 
\new{branch} is used to generate appearance feature for data association.}\label{fig::pipeline}
\vspace{-0.25in}
\end{figure*}

\subsection{Counting, detection and tracking using density maps}

Object counting aims to estimate the number of objects in an image, and it is different from object detection, which focuses on the individual object and usually suffers from occlusion. Crowd density map is an effective method for object counting, where the object count in a region corresponds to the sum over that region.
Recent approaches use deep learning techniques to learn crowd density map~
and have proved that it is helpful to locate objects in crowd scenes.


Using crowd density map, \cite{rodriguez2011density} primarily proposes a ``density-aware'' detection and tracking framework, where the detections are encouraged to be consistent with the crowd density map. However, it only adopts a nearest-neighbour strategy for tracking and doesn't model object appearance, which could fail when two objects stay close to each other. 
As a pioneer work, \cite{ma2015small} adopts crowd density map for small object detection, where integer programming is used to recover object locations from sliding windows over the density map. \cite{lian2019density}~proposes a depth-adaptive method to simultaneously estimate head counts and locate head positions. The method uses two independent branches to predict density map and head locations separately, but it doesn't establish a connection between the two tasks. 
\cite{sam2020locate}~introduces a multi-column network for crowd counting, and head location is obtained by a classification task of the predefined boxes.
In~\cite{ren2018fusing}, a fusion tracker is proposed by combing crowd density map and a visual object tracker for tracking in crowd scenes. This method only considers single object tracking, and is not applicable for multi-object tracking. \cite{wan2021body}~proposes a joint body-face detector for multi-object tracking, but it may generate miss-matching between bodies and faces in crowd scenes. \cite{ren2020tracking}~presents a novel tracking paradigm, where detection, counting, and tracking are jointly formulated as a network flow problem on crowd density maps. Though this method can achieve detection and tracking through a global optimization, it should take much time to create a network-flow graph. 

In contrast to the existing methods, our model jointly formulates counting, object detection and appearance embedding using a multi-task learning scheme.  By imposing mutual object-count constraints between object detection and crowd density map, the two tasks can be simultaneously optimized and enhanced, making the tracking task robust to crowd scenes. Besides, the joint model can perform real-time detection and tracking simultaneously.

\section{Joint Counting, Detection and reID model}
\label{text:tbc}
Our CountingMOT model has three homogeneous branches, and it optimizes object detection, counting and reID feature
simultaneously in one framework, shown in Fig.~\ref{fig::pipeline}. Different from other \emph{joint detection-and-tracking} methods, our model uses counting task (i.e., crowd density map) to enhance detection task in crowd scenes.
The workflow of our method is as follows. Multi-scale features of an input image are first extracted by an encoder-decoder backbone, 
which are then adopted to simultaneously generate object detections, density map and reID features. For the object detection branch, it is not only supervised by detection loss, but also encouraged to be consistent with the crowd density map. Meanwhile, the predicted density map is constrained by the object count from object detections at the same time. The mutual constraints between detection and crowd density map can enhance each other, and work well for multi-object tracking in crowd scenes. Together with object detections, the reID features are finally used for data association. In the remainder of this section, we will introduce each component of our model: feature extractor, detection task, density map estimation task and reID task respectively.

\subsection{Feature Extractor}
Following~\cite{zhang2021fairmot}, we also use DLA (Deep Layer Aggregation) as the backbone to extract multi-scale features. DLA is preliminary proposed for image classification, and then is enhanced by~\cite{zhou2019objects} with hierarchical skip connections. To improve feature map resolution, DLA~iteratively aggregates low-resolution features to high-resolution features. Besides, the original convolution at each upsampling layer is replaced with deformable convolution to dynamically perceive different object scales. The revised DLA aggregates semantic and spatial fusion to capture information across layers, which has been proven more effective for object detection and tracking~\cite{zhou2019objects,zhou2020tracking, zhang2021fairmot}. The parameter setting in DLA of this work is the same with the previous works.  \rwh{In the experimental part, we also prove that the CountingMOT can be built on 
YOLOX~\cite{ge2021yolox} detector.}

As shown in Fig.~\ref{fig::pipeline}, for an input image with size $H_\text{in} \times W_\text{in}\times 3$, the output feature map of the feature extractor is of size $H\times W \times 256$. Here, $H = H_\text{in}/R$, $W = W_\text{in}/R$, where the downsampling factor $R$ is set to 4 in this work.

After feature extraction, we add three homogeneous branches to predict object detection, density map and reID feature, respectively. Each branch has a $3 \times 3$ convolutional layer with $256$ channels after the feature extractor, followed by a $1 \times 1$ convolutional layer to produce the final output.

\subsection{Detection task}
The detection 
\new{branch} is an anchor-free method, responsible for estimating the object centers. The anchor-free method locates an object through object center, which is convenient for re-Identification feature learning~\cite{zhang2021fairmot}.
It produces a set of object detections $\{\left(\hat{\textbf{p}}_1,\hat{\textbf{s}}_1 \right), \left(\hat{\textbf{p}}_2, \hat{\textbf{s}}_2\right), ... \}$ for each class $c\in \left\{ 0,...,C-1 \right\}$, where~${{\hat{\textbf{p}}}_{i} = \left(\hat{p}_i^x, \hat{p}_i^y\right)}$ is the object center and ${{\hat{\textbf{s}}}_{i} = \left(\hat{h}_i, \hat{w}_i\right)}$ indicates the object scale (height and width). Specifically, the object centers are produced by a heatmap~$\hat{M}\in {{\left[ 0,1 \right]}^{{{{H}}}\times {{{W}}}\times C}}$, and the object scale is also generated by a scale map~$\hat{S}\in {{\left[ 0,1 \right]}^{{{{H}}}\times {{{W}}}\times 2}}$. Thus, the final output of the detection 
\new{branch} has $C + 2$ channels.

Given a set of GT (Ground Truth) annotated boxes $\left\{ \left( {{{\textbf{p}}}_{i}},{{{\textbf{s}}}_{i}} \right) \right\}_{1}^{N}$, we first generate a GT heat map~${M}\in {{\left[ 0,1 \right]}^{{{{H}}}\times {{{W}}}\times C}}$ using a Gaussian kernel
\begin{equation} \label{eq1}
	M=\sum\limits_{i=1}^{N}{\exp -\frac{{{\left( x-p_{i}^{x} \right)}^{2}}+{{\left( y-p_{i}^{y} \right)}^{2}}}{2\sigma _{i}^{2}}},
\end{equation}
where $N$ represents the number of objects in the image, $(x, y)$  represents a location at the heat map, and $\sigma_i$ is the standard deviation, determined by the object scale. Then, the loss function for object center is a pixel-wise regression with a focal loss~\cite{lin2017focal}

\begin{equation}\label{eq2}
\resizebox{0.44\textwidth}{!}{$
{\mathcal{L}_{\text{center}}}=-\frac{1}{N}\sum\limits_{xyc}{\left\{ \begin{aligned}
  & {{\left( 1-{{{\hat{M}}}_{xyc}} \right)}^{\alpha }}\text{log}\left( {\hat{M}_{xyc}} \right)\text{ ~~~~~~~~~~~~~~~if }{{M}_{xyc}}=1 \\ 
 & {{\left( 1-{{M}_{xyc}} \right)}^{\beta }}{{\left( {{{\hat{M}}}_{xyc}} \right)}^{\alpha }}\text{log}\left( 1-{\hat{M}_{xyc}} \right)\text{ otherwise} \\ 
\end{aligned} \right.}$},
\end{equation}
where $\alpha$ and $\beta$ are the hyper-parameters in focal loss, and are set to $2$ and $4$, respectively.

For scale estimation, we can directly use $L_1$ loss between ${{{\hat{\textbf{s}}}}^{i}}$ and ${{\textbf{s}}^{i}}$. Since the downsampling factor $R$ in the final feature map introduces quantization errors for object center, we thus \new{add} an additional offset branch to compensate for the errors. The GT offset is obtained by ${{\textbf{o}}^{i}}=\left( \frac{p_{i}^{x}}{R},\frac{p_{i}^{y}}{R} \right)-\left( \left\lfloor \frac{p_{i}^{x}}{R},\frac{p_{i}^{y}}{R} \right\rfloor  \right)$, and the corresponding predicted one is represented as ${\hat{\textbf{o}}^{i}}$. Then, the loss for scale and offset can be written as
\begin{equation}\label{eq3}
{\mathcal{L}_{\text{scale}}}=\sum\limits_{i=1}^{N}{\left\| {{{\hat{\textbf{s}}}}_{i}}-{{\textbf{s}}_{i}} \right\|+{{\left\| {{{\hat{\textbf{o}}}}_{i}}-{{\textbf{o}}_{i}} \right\|}_{1}}}.
\end{equation}

Overall, the total loss for object detection task is
\begin{equation}\label{eq4}
\mathcal{L}_{\text{det}} = \mathcal{L}_{\text{center}} + \mathcal{L}_{\text{scale}}.
\end{equation}

\subsection{Density map estimation task}
\new{Crowd density map estimation was first proposed for object counting~\cite{lempitsky2010learning}, where the sum over a region in the map corresponds to the object count in that region. It can provide an informative clue for object localization since the GT density map is usually generated by blurring object annotations with a Gaussian kernel.} 
\new{In this work, for density map estimation, we directly} use the extracted features from the backbone to generate a density map $\hat{D}$ with size~${{{{H}}}\times {{{W}} \times 1}}$. Following the typical \new{density map generation methods~\cite{li2018csrnet,song2021rethinking,wang2021neuron}}, 
we adopt a scale-adaptive strategy to generate density map for crowd scenes. All the object centers in an image are first represented by an indicator matrix, where ``1" corresponds to an object while ``0" is the background. Then, the GT density map is obtained by blurring each object center ${{\textbf{p}}_{i}}$ using a Gaussian kernel
\begin{equation}\label{eq5}
D\left( \textbf{p} \right)=\sum\limits_{j=1}^{N}{\delta \left( \textbf{p}-{{\textbf{p}}_{j}} \right)\otimes {{G}_{\sigma _{j}}}\left(\textbf{p} \right)},
\end{equation}
where $\delta$ is the delta function, $\otimes$ indicates the convolution operator, and ${{G}_{\sigma _{j}}}$ is the 2D Gaussian kernel with standard deviation ${\sigma _{j}}$. Here, ${\sigma _{j}} = \gamma {{\bar{d}}_{j}}$ and ${{\bar{d}}_{j}}$ is the average distances of $k$ nearest neighbours. Different from the object heat map, the Gaussian kernel ${{G}_{\sigma _{j}}}$ is normalized to ``1'' to keep the sum over the density map consistent with the number of objects. Finally, the Mean
Square Error (MSE) loss and the Structural Similarity Index (SSIM)~\cite{wang2004image} 
loss are jointly adopted to measure the difference between $\hat{D}$ and ${D}$
\begin{equation}\label{eq6}
{{\mathcal{L}}_{\text{cnt}}}={\left\| \hat{D}-\mu \cdot D \right\|_{2}^{2}} + \text{SSIM}\left(\hat{D},  ~\mu\cdot D\right),
\end{equation}
where $\mu$ is an amplification factor to accelerate convergence and lower estimation error~\cite{gao2019c}.

\subsection{Mutual constraints between detection and counting}
Object detection has achieved significant progress in recent years~\cite{zou2019object}, 
but it usually fails to handle crowd scenes due to the heavy occlusions. Crowd density map is designed for crowd counting, and can provide an informative clue for object detection. By exploiting object count  constraint, we establish connections between object detection and density map estimation, to jointly improve the tracking performance.

\subsubsection{Detection with counting}
After obtaining object detections from the detection 
\new{branch}, we first generate an indicator matrix $U$ to represent the candidate object centers. Similar to GT density map generation (see~(\ref{eq5})), the candidate detection matrix $U$ is then blurred by a Gaussian kernel to generate a density map $U\otimes {{G}_{\sigma _{j}}}$, where ${\sigma _{j}}$ is also determined by its $k$ nearest neighbours. Intuitively, the density map $U\otimes {{G}_{\sigma _{j}}}$ should be consistent with the density map estimated by the density 
\new{branch} as shown in Fig.~\ref{fig::pipeline}. Thus, the additional constraint for detection task can be formulated as
\begin{equation}\label{eq7}
{{\mathcal{L}}_{\text{dc}}}={\left\| U\otimes {{G}_{\sigma _{j}}} - \hat{D}\right\|_{2}^{2}}.
\end{equation}
Usually, the estimated density map $\hat{D}$ can accurately provide the object count information in a region, and thus it can implicitly help detection task to find occluded or missed detections. Note that $U$ contains the trainable variables, and $\hat{D}$ is fixed in this constraint.

\subsubsection{Counting with detection}
Detection task should accurately locate objects, and thus it usually has few false detections. \new{Crowd density map focuses on counting of the whole scene, 
and we can use object detection task to 
enhance the localization ability of density map estimation task.} For the predicted density map $\hat{D}$, we first predefine a set of sliding windows which move vertically and horizontally in the density map. Each 2D sliding window \new{is vectorized as a 1D mask vector $\textbf{w} \in {{\left\{ 0,1 \right\}}^{HW}}$}, where ``1" means that a pixel is within the sliding window, and ``0" otherwise. Thus, for a specific sliding window $\textbf{w}_k$, its object count can be obtained from density map $\hat{D}$
\begin{equation}\label{eq8}
{{n}_{k}}={{\left( {\textbf{w}_{k}} \right)}^{T}}\hat{\textbf{d}},
\end{equation}
where $\hat{\textbf{d}}$ is \new{also} the vectorization of the density map $\hat{D}$. Also, the object count can be computed from the object detections
\begin{equation}\label{eq9}
{{n}_{k}}={{\left( {\textbf{w}_{k}} \right)}^{T}}\textbf{u},
\end{equation}
where $\textbf{u}$ is the vectorization of object detections $U$. The predicted density map thus can be optimized through minimizing the counting difference between $(\ref{eq8})$ and $(\ref{eq9})$
\begin{equation}\label{eq10}
{{\mathcal{L}}_{\text{cd}}}=\frac{1}{K}\sum\limits_{k=1}^{K}{{{\left( {\textbf{w}^{T}_k}\hat{\textbf{d}}-{\textbf{w}^{T}_k}\textbf{u} \right)}^{2}}},
\end{equation}
where $K$ is the number of sliding windows, $\hat{\textbf{d}}$ is the training variable and $\textbf{u}$ is fixed. \new{ The sliding windows are densely sampled from the crowd density map $\hat{D}$ with size ${{{{H}}}\times {{{W}} \times 1}}$, which means that the number of sliding windows $K$ equals to the product of $H$ and $W$.} 
The sliding window $\textbf{w}_{k}$ is an important factor for object detection, and further affects the tracking performance. 
The size of $\textbf{w}_k$ means how much region is used to calculate the difference between density map and object detections.
Usually, it is set to the average size of the objects in a scene. For a small $\textbf{w}_k$, the sum (object count) within the crowd density map is less than 1, which may cause missed detections in the detection task due to the mutual constraints. Also, for a large $\textbf{w}_k$, it contains too many objects, resulting in inaccurate detections. Please refer to section~\ref{sec:IV-D} for further analysis.

\subsection{ReID task}
ReID task tries to learn appearance features to distinguish objects. Similar to object heat map $\hat{M}$ and crowd density map $\hat{D}$, the reID branch outputs a feature map $\hat{E}\in {{\mathbb{R}}^{{{{H}}}\times {{{W}}}\times 128}}$, where $\hat{E}_{ij}$ represents the embedding feature centered at $\left(i, j \right)$, and $128$ is the feature dimension.
Following the work~\cite{zhang2021fairmot}, the reID task can be regarded as a classification problem. For a given GT box, 
its embedding feature can be first extracted from the feature map $\hat{E}$, and then is converted to a class distribution vector $\textbf{q}^i$ through a softmax loss. The GT class label for the box can be represented as a one-hot vector, denoted as $\textbf{v}^i$. The reID loss then can be computed as
\begin{equation}\label{eq11}
{{\mathcal{L}}_{\text{id}}}=-\sum\limits_{i=1}^{N}{\sum\limits_{l=1}^{L}{{\textbf{v}^{i}}\left( l \right)\text{log}\left( \textbf{q}^i\left( l \right) \right)}},
\end{equation}
where $L$ is the number of instance identities, and $N$ also denotes the number of objects in the image. For training, only the embedding features located at object centers are used. To improve the robustness of reID features, we adopt image transformations \new{including} HSV~(Hue, Saturation, Value) augmentation, rotation, scaling, translation and shearing for data preparation.

\subsection{Overall loss for training CountingMOT}
For CountingMOT, the object detection, density map estimation and reID feature can be trained simultaneously by using uncertainty weight~\cite{kendall2018multi}
\begin{equation}\label{eq12}
\begin{aligned}
  & {{\mathcal{L}}_{\det \text{-dc}}}={{\mathcal{L}}_{\text{det}}}+{{\mathcal{L}}_{\text{dc}}} \\ 
 & {{\mathcal{L}}_{\text{cnt-cd}}}={{\mathcal{L}}_\text{cnt}}+{{\mathcal{L}}_{\text{cd}}} \\ 
\end{aligned},
\end{equation}
\begin{equation}\label{eq:all}
\begin{aligned}
   {{\mathcal{L}}_{\text{total}}}=\frac{1}{2}&\left( \frac{1}{{{e}^{w1}}}{{\mathcal{L}}_{\det \text{-dc}}}+\frac{1}{{{e}^{w2}}}{{\mathcal{L}}_{\text{cnt-cd}}}+\frac{1}{{{e}^{w3}}}{{\mathcal{L}}_{\text{id}}} \right) \\ 
 & +\left( w1+w2+w3 \right) \\ 
\end{aligned},
\end{equation}
where $w_1$, $w_2$ and $w_3$ \new{are} trainable parameters to automatically balance the three tasks. For training, we first generate heat map, size map, box offset, density map and one-hot representation for each object identity. Then, the total loss is computed between the GT labels and the predicted outputs.

\setlength{\tabcolsep}{8pt}
\begin{table*}[!htb]
\small
\begin{center}
\caption{Comparison with the state-of-the-art trackers under the ``private detector'' protocol. The two-stage trackers are labeled by ``*''. The best results of each dataset are shown in {\bf bold}, and the second best are in {\underline {underline}}.}
\label{table:sota}
\begin{tabular}{llccccccccc}
\toprule
Dataset & Tracker & MOTA$\uparrow$ & IDF1$\uparrow$&HOTA$\uparrow$ & MT$\uparrow$ & ML$\downarrow$ &\new{FP$\downarrow$}&\new{FN$\downarrow$}& IDs$\downarrow$ & FPS$\uparrow$\\
\midrule
MOT16 
 & DeepSORT~\_2\textsuperscript{*}~\cite{wojke2017simple} & 61.4 & 62.2 &50.1& 32.8\% & 18.2\%&12,852&56,668 & 781 &  6.4\\

&TLR~\cite{wang2021multiple}&\underline{76.6}&74.3&61.0&47.8\%&{\bf 13.3\%}&10,860&\underline{30,756}&979&15.9 \\

&Trackor++~\cite{bergmann2019tracking}&54.4&52.5&42.3&19.0\%&36.9\%&\textbf{3,280}&	79,149&{682}&1.5 \\

&GSDT~\cite{wang2021joint}&74.5&68.1&56.6&41.2\%&17.3\%&8,913&36,428&1,229& 1.6 \\
&QuasiDense~\cite{pang2021quasi}&69.8&67.1&54.5&41.6\%&19.8\%&9,861	&44,050&1,097&20.3 \\
&TraDeS~\cite{wu2021track}&70.1&64.7&53.2&37.3\%&20.0\%&\underline{8,091}&45,210&1,144&22.3 \\
&TubeTK~\cite{pang2020tubetk}&66.9&62.2&50.8&39.0\%&16.1\%&11,544&47,502&1,236&1.0\\
&FairMOT~\cite{zhang2021fairmot} & {75.7} & \underline{75.3} &61.6&  {48.1\%} & \underline{14.4\%}& 13,501 &41,653 & \underline{621} & {\bf 25.9}\\

&GRTU\textsuperscript{*}~\cite{wang2021general}
&76.5&{\bf 75.9}&{\bf 62.6}&{\bf 51.5}\%&17.0\%&11,438&30,866&{\bf 584}&0.3 \\
&CTrackerV1~\cite{peng2020chained}
&67.6&57.2&48.8&32.9\%&23.1\%&8,934&48,350&1,897&6.8 \\
&MeMOT~\cite{cai2022memot}&69.7&72.6&57.4&44.9\%&16.6\%&14,595&34,595&845& -- \\
& CountingMOT (ours)&{\bf 77.6}&75.2&\underline{62.0}&\underline{50.7\%} & {14.8\%}&12,337&\textbf{27,382}	&1,087&\underline{24.9} \\
\cline{2-11}
\rowcolor{lightgray}
\cellcolor{white}& UTM~\cite{you2023utm}&{ 81.1}&79.0&{64.1}&{58.2\%} & {12.6\%}&11,722&{22,367}&440&{451.8} \\
\rowcolor{lightgray}
\cellcolor{white}& CountingMOT\_v2 (ours)&{ 79.7}&78.5&{63.6}&{65.5\%} & {9.0\%}&17,158&{18,195}&1,643&{26.4} \\
\midrule
MOT17 
&TBC~\cite{ren2020tracking}&53.9 & 50.0& 40.9&20.2\%&36.7\%&24,584&232,670	&4,612&6.7 \\
&SST\textsuperscript{*}~\cite{sun2019deep} & 52.4 & 49.5 &39.3& 21.4\% & 30.7\%&25,423& 23,4592& 8,431 & 3.9\\
&TLR~\cite{wang2021multiple}&\underline{76.5}&73.6&60.7&47.6\%&{\bf 12.7\%}&29,808&\underline{99,510}&3,369&15.6 \\

&Trackor++~\cite{bergmann2019tracking}&56.3&55.1&44.8&21.1\%&35.3\%&\textbf{8,866}&	235,449&\underline{1,987}&1.5 \\

&GSDT~\cite{wang2021joint}&66.2&68.7&55.5&40.8\%&18.3\%&26,397&120,666&3,318& 4.9 \\
&QuasiDense~\cite{pang2021quasi}&68.7&66.3&53.9&40.6\%&21.9\%&26,589&	146,643&3,378&20.3 \\
&TraDeS~\cite{wu2021track}&69.1&63.9&52.7&37.3\%&20.0\%&20,892&150,060	&3,555&22.3 \\
&TubeTK~\cite{pang2020tubetk} & 63.0 & 58.6 &48.0& 31.2\% & 19.9\%&27,060&	177,483 & 4,137 & 3.0\\
&CenterTrack~\cite{zhou2020tracking} & 67.8 & 64.7 &52.2 & 36.4\% & 21.5\%&\underline{18,498}&	160,332 & {2,583} & 17.5\\
& FairMOT~\cite{zhang2021fairmot}& {73.7} & 72.3& {59.3} & {43.2\%} & {17.3\%}& 27,507	&117,477 & 3,303 & {\bf 25.9}\\
&GRTU\textsuperscript{*}~\cite{wang2021general}
&74.9&{\bf 75.0}&{\bf 62.0}&\underline{49.7}\%&18.9\%&32,007&107,616&{\bf 1,812}&3.6 \\
& CTrackerV1~\cite{peng2020chained}
& 66.6 & 57.4 & 49.0 &32.2\% & 24.2\%&22,284&160,491 & 5,529 & 6.8\\
&PermaTrack~\cite{tokmakov2021learning}
&73.1&67.2&54.2&42.3\%&19.1\%&28,998&115,104&3,571 &11.9 \\
& CSTrack~\cite{liang2022rethinking}
&74.9&{72.3}&41.5&50.4\%&15.5\%&23,847&114,303&3,196&4.5 \\
&MeMOT~\cite{cai2022memot}&69.0&72.5&56.9&43.8\%&18.0\%&37,221&115,248&2,724&--\\
&Trackformer~\cite{meinhardt2022trackformer} &74.1&68.0&57.3&47.3\%&10.4\%&34,602&	108,777&2,829&5.7\\

&CountingMOT (ours)&{\bf 78.0}&\underline{74.8}&\underline{61.7}&{\bf 49.8\%} & \underline{15.4\%}&28,233&\textbf{92,247}&3,453&\underline{24.9} \\

\cline{2-11}

\rowcolor{lightgray}
\cellcolor{white}& Motion\_Track\textsuperscript{*}~\cite{qin2023motiontrack}&{81.1}&\underline{80.1}&\underline{65.1}&{55.5\%} & {16.7\%}&{\bf 23,802}&{81,660}&{\bf 1,140}&{15.7} \\

\rowcolor{lightgray}
\cellcolor{white}& SUSHI\textsuperscript{*}~\cite{cetintas2023unifying}&{81.1}&{\bf 83.1}&{\bf 66.5}&\underline{58.3\%} & \underline{13.2\%}&32,313&\underline{73,245}&\underline{1,149}&{21.1} \\

\rowcolor{lightgray}
\cellcolor{white}& ByteTrack\textsuperscript{*}~\cite{zhang2022bytetrack}&{80.3}&77.3&{63.1}&{53.2\%} & {14.5\%}&25,491&{83,721}&2,196&\underline{29.6} \\

\rowcolor{lightgray}
\cellcolor{white}& UTM~\cite{you2023utm}&{\bf 81.8}&78.7&{64.0}&{56.4\%} & {13.8\%}&\underline{25,077}&{76,298}&1,431&{\bf 1,355.5} \\
 
 \rowcolor{lightgray}
\cellcolor{white}& CountingMOT\_v2 (ours)&\underline{81.3}&78.4&{63.6}&{\bf 63.2\%} & {\bf 9.9\%}&38,565&{\bf 61,968}&5,118&{26.4} \\

\midrule
MOT20 
& TBC~\cite{ren2020tracking}&54.4&50.1&--&33.4\%&19.7\%&37,937&195,242&2,580&5.6 \\
&Trackor++~\cite{bergmann2019tracking}&52.6&52.7&42.1&29.4\%&26.7\%&\textbf{6,930}&	236,680&{\bf 1,648}&1.2 \\
& FairMOT~\cite{zhang2021fairmot}& \underline{61.8} & {67.3} & \underline{54.6}& {\bf 68.8\%} & {\bf 7.6\%}&103,440&\textbf{88,901}&{5,243} & {\bf 13.2}\\
&GSDT~\cite{wang2021joint}&67.1&68.6&53.6&53.1\%&13.2\%&31,913&135,409&3,230& 1.5 \\

&MLT\textsuperscript{*}~\cite{zhang2020multiplex}&48.9&54.6&43.2&30.9\%&22.1\%&45,660&	216,803&{2,187}&3.7 \\

&TransCenter~\cite{xu2021transcenter}
&58.5&49.6&43.5&48.6\%&14.9\%&64,217&146,019&4,695&1.0 \\
& CSTrack~\cite{liang2022rethinking}
&66.6&\underline{68.6}&54.0&50.4\%&15.5\%&25,404&144,358&3,196&4.5 \\
&MeMOT~\cite{cai2022memot}&66.1&63.7&54.1&57.5\%&14.3\%&47,882 &137,983&\underline{1,938}&--\\
&Trackformer~\cite{meinhardt2022trackformer}&68.6&65.7&54.7&53.6\%&
14.6\%&\underline{20,348}&140,373&2,474&5.7\\

& CountingMOT (ours)&{\bf 70.2}&{\bf 72.4}&{\bf 57.0}&\underline{62.0\%} & \underline{12.1\%}&33,531&\underline{117,886}&{2,795}&\underline{12.6} \\

\cline{2-11}

\rowcolor{lightgray}
\cellcolor{white}& Motion\_Track\textsuperscript{*}~\cite{qin2023motiontrack}&{78.0}&76.5&{62.8}&{\bf 71.3\%} & {9.5\%}&28,629&{84,152}&\underline{1,165}&{15.0} \\

\rowcolor{lightgray}
\cellcolor{white}& SUSHI\textsuperscript{*}~\cite{cetintas2023unifying}&{74.3}&{\bf 79.8}&{\bf 66.5}&{63.0\%} & {14.5\%}&{\bf 16,841}&{115,462}&{\bf 706}&{5.3} \\

\rowcolor{lightgray}
\cellcolor{white}& ByteTrack\textsuperscript{*}~\cite{zhang2022bytetrack}&{77.8}&75.2&{61.3}&{69.2\%} & {9.5\%}&26,249&{87,594	}&1,223&\underline{17.5} \\

\rowcolor{lightgray}
\cellcolor{white}& UTM~\cite{you2023utm}&\underline{78.2}&76.9&{62.5}&\underline{70.4\%} & {\bf 8.6\%}&29,964&{\bf 81,516	}&1,228&{\bf 722.4} \\
 
 \rowcolor{lightgray}
\cellcolor{white}& CountingMOT\_v2 (ours)&{\bf 78.9}&\underline{78.6}&\underline{63.6}&{70.0\%} & \underline{8.9\%}&\underline{24,368}&\underline{83,437}&1,232&{14.4} \\

\bottomrule
\end{tabular}
\end{center}
\vspace{-0.2in}
\end{table*}

\subsection{Online multi-object tracking}
For each frame in a video sequence, the CountingMOT model jointly outputs object detections and their corresponding reID features. To avoid interference of object distractors, we adopt the hierarchical strategy for data association. 
First, the Kalman Filter is used to predict the location of each previous tracklet in the current frame. Then, the cosine distance on reID features and Mahalanobis distance on bounding boxes are computed, respectively. \new{Through} a weighting parameter $\lambda$ ($\lambda = 0.98$), a final distance matrix can be obtained which is applied for preliminary matching. For the unmatched tracklets and detections, we further match them using IoU~(Intersection over Union) with a given threshold $\tau$ ($\tau = 0.5$). After that, the final unmatched tracklets will be maintained for 30 frames  before termination, while the unmatched detections are initialized as new trackers. More details can be found in~\cite{wojke2017simple}.

\section{Experiments}
\label{text:experiments}
In this section, we evaluate the CountingMOT model on some public MOT datasets~\emph{MOT16}, \emph{MOT17} and \emph{MOT20}. Note that \emph{MOT20} contains some extremely crowd scenes which \new{are} very suitable to validate the anti-occlusion ability of MOT trackers. Also, we perform ablation studies to prove the effectiveness of the mutual constraints between detection and counting. 

\subsection{Datasets and metrics} \label{sec::datasts}
Following~\cite{zhang2021fairmot}, we also adopt the same experimental setup to train our CountingMOT model. Six additional datasets (some with box and identity annotations) are used for pre-training, and they contain various pedestrian scenes (e.g., street, station, malls or the wild). 

For evaluation, the commonly CLEAR MOT metrics are adopted~\cite{bernardin2008evaluating}, 
e.g., MOTA (Multiple Object Tracking Accuracy), IDS (ID Switch), MT (Mostly Tracked Trajectories), ML (Mostly Lost Trajectories), \new{FP (Number of False Positives), FN (Number of False Negatives)} and etc. Also, IDF1~(ID F1 Score) is used to measure the correctly identified detections~\cite{ristani2016performance}. 
Recent approach~\cite{luiten2021hota} thinks that previous metrics overemphasize either detection or association, and thus it proposes a new MOT metric HOTA (Higher Order Tracking Accuracy) which matches at the detection level while considering association over the whole trajectory.

\subsection{Implementation details}
For training, we use the pre-trained parameters on the COCO dataset to initialize the CountingMOT model. The entire network is trained 
using the Pytorch framework with a batch size 12. The Adam optimizer is used for training with 30 epochs, where the starting learning rate is $1e^{-4}$ and then decays to  $1e^{-5}$ at the 20 epochs. The input image for the model is resized to $1088 \times 608$ (namely $W_\text{in} = 1088$, $H_\text{in} = 608$), and the standard transforms including color jittering, rotation and scaling are also adopted. In total, the whole training process takes about 10 hours.

For the hyper-parameters, $\mu$ in (\ref{eq6}) is set to $1000$ for fast convergence. The size of the sliding window $\textbf{w}_k$ in (\ref{eq8}) is set to $19 \times 19$. We do some ablation studies in the experimental part to analyze the effect of the two parameters. 

\subsection{Evaluation on MOT challenge}
We evaluate our CountingMOT model on the datasets from the \emph{MOT challenge}, and compare it with the recent state-of-the-art approaches including two-stage ones (e.g., \cite{sun2019deep, wang2021general}) and JDT ones (e.g., \cite{zhou2020tracking,zhang2021fairmot,cai2022memot, meinhardt2022trackformer}). The tracking results on \emph{MOT16}, \emph{MOT17} and \emph{MOT20} are summarized in Tab.~\ref{table:sota} (the white part of the table). Note that all the results are directly taken from the related papers or MOT leaderboard. 

For \emph{MOT17}, our CountingMOT tracker performs the best among all the methods in terms of MOTA. The two-stage method GRTU~\cite{wang2021general} has a higher HOTA, since it adopts a recurrent module to associate potential tracks through long-term dependency. However, this method runs a little slowly (\new{3.6 FPS}), and can't be applied to real-time tracking. By introducing density map as auxiliary information, our CountingMOT model can significantly improve FairMOT~\cite{zhang2021fairmot} (e.g., MOTA can be improved from 73.7 to 78.0). \new{Besides, CountingMOT has less FP and FN compared with FairMOT on \emph{MOT16} (FP of 12,337 vs 13501, FN of 27,382 vs 41,653), which implies that crowd density map can help to find missed object detections. For \emph{MOT17}, CountingMOT can also significantly improve FairMOT by decreasing FN from $117,477$ to $92,247$, but it slightly introduces more FP compared with FairMOT ($28,233$ vs $27,507$). }

\begin{figure*}[!hbtp]
\centering
\captionsetup[subfigure]{labelformat=empty}

\vspace{-0.10in}
\subfloat{\includegraphics[width=\widththird]{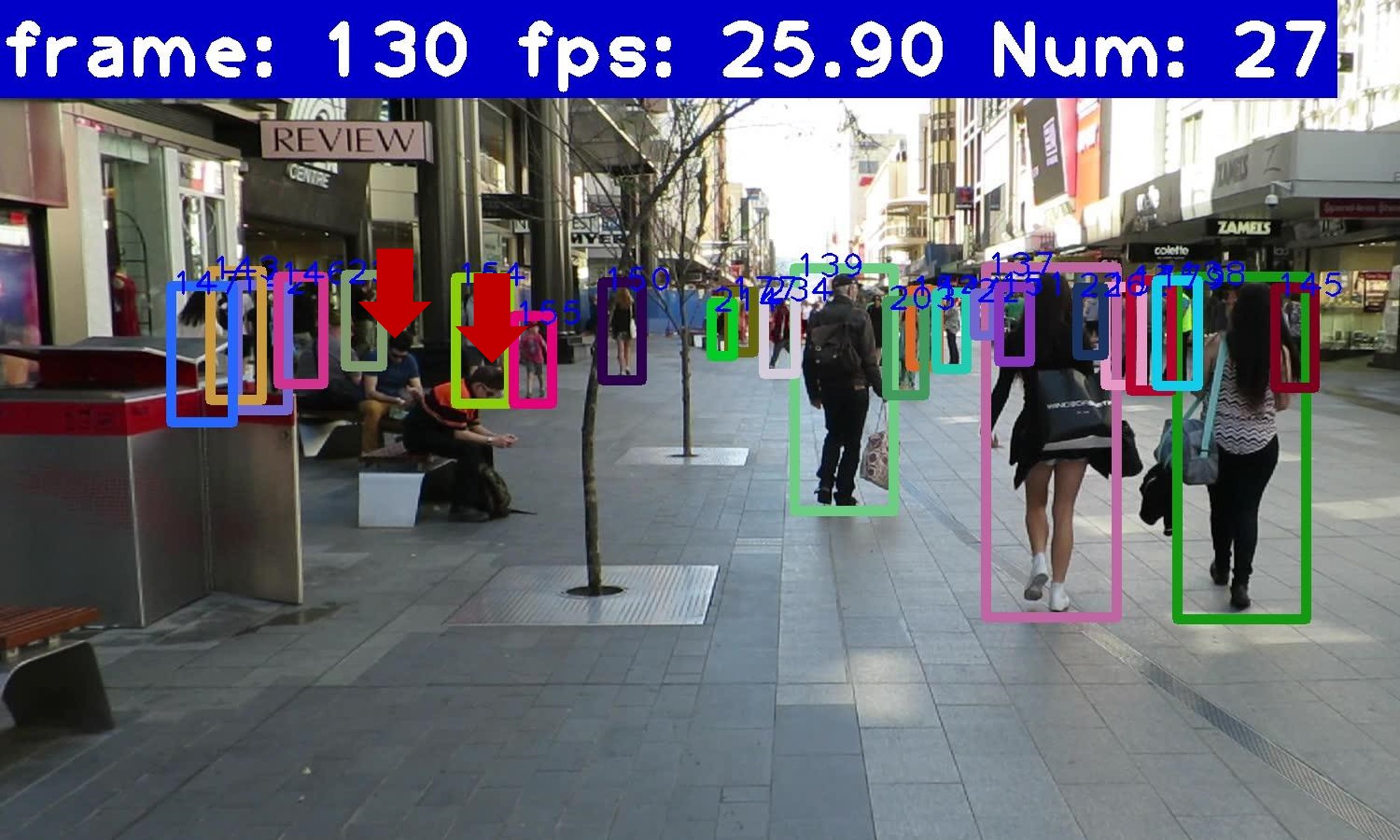}} \hspacefigure
\subfloat{\includegraphics[width=\widththird]{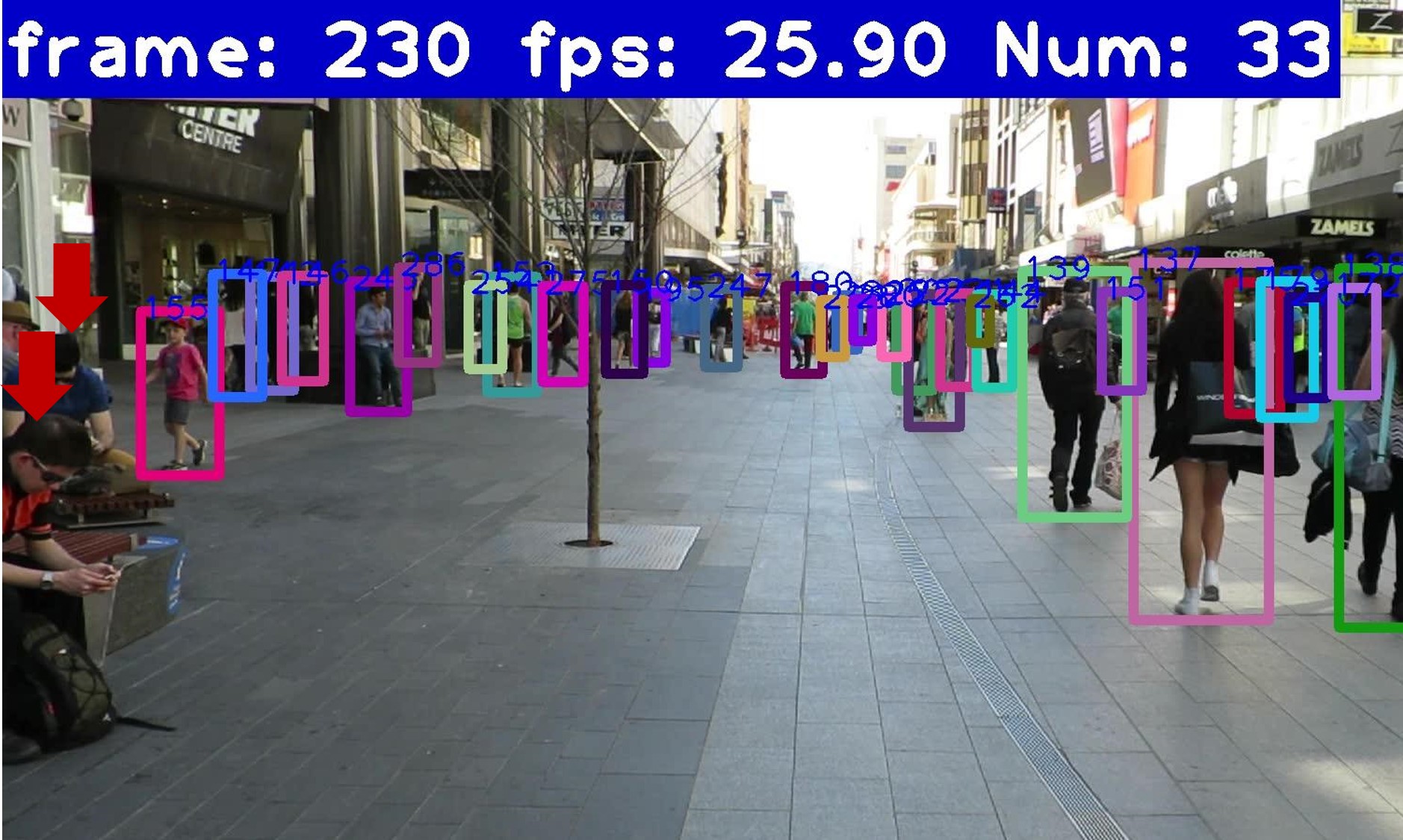}} \hspacefigure
\subfloat{\includegraphics[width=\widththird]{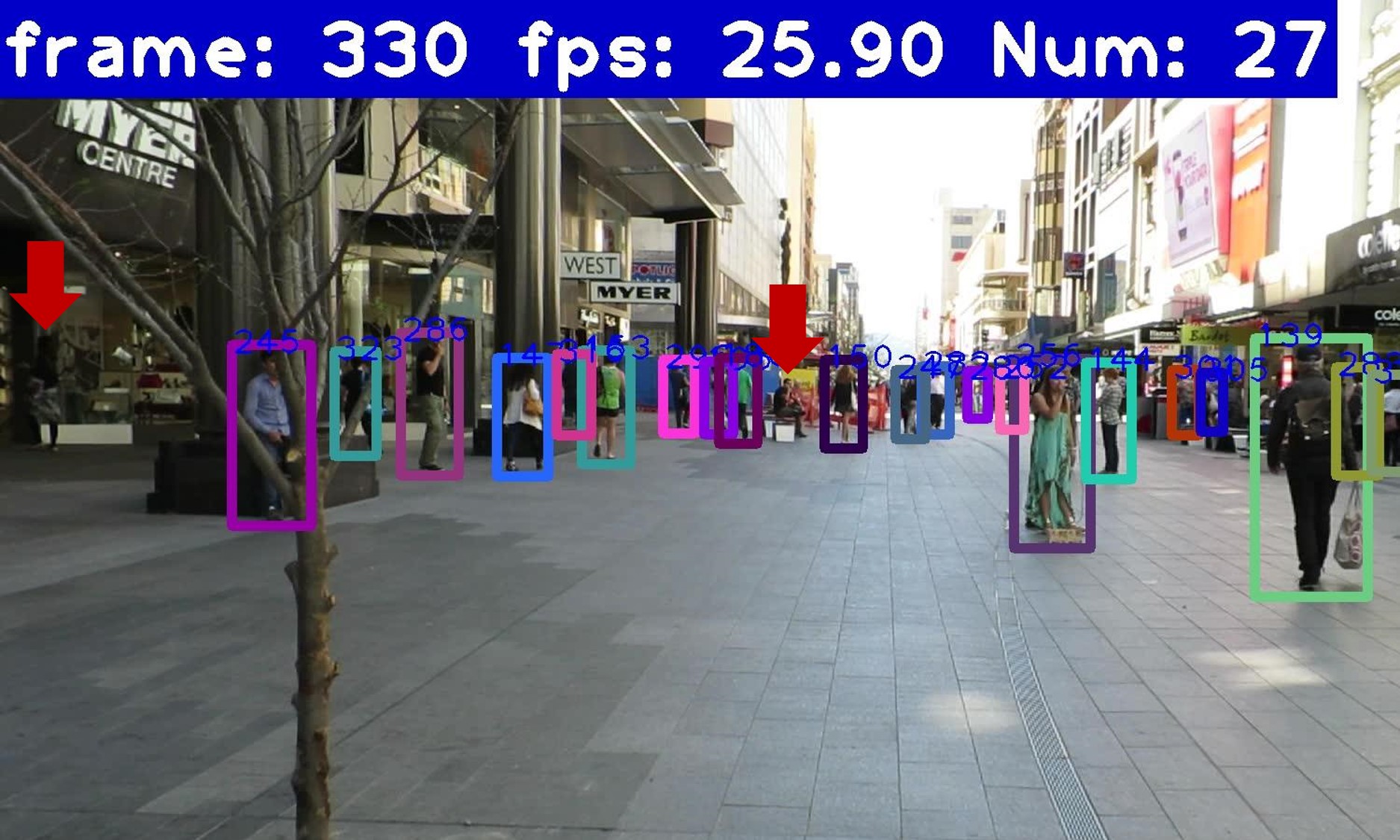}} \hspacefigure
{\small  FairMOT~\cite{zhang2021fairmot}~(IJCV 2021)} \\

\vspace{-0.10in}
\subfloat{\includegraphics[width=\widththird]{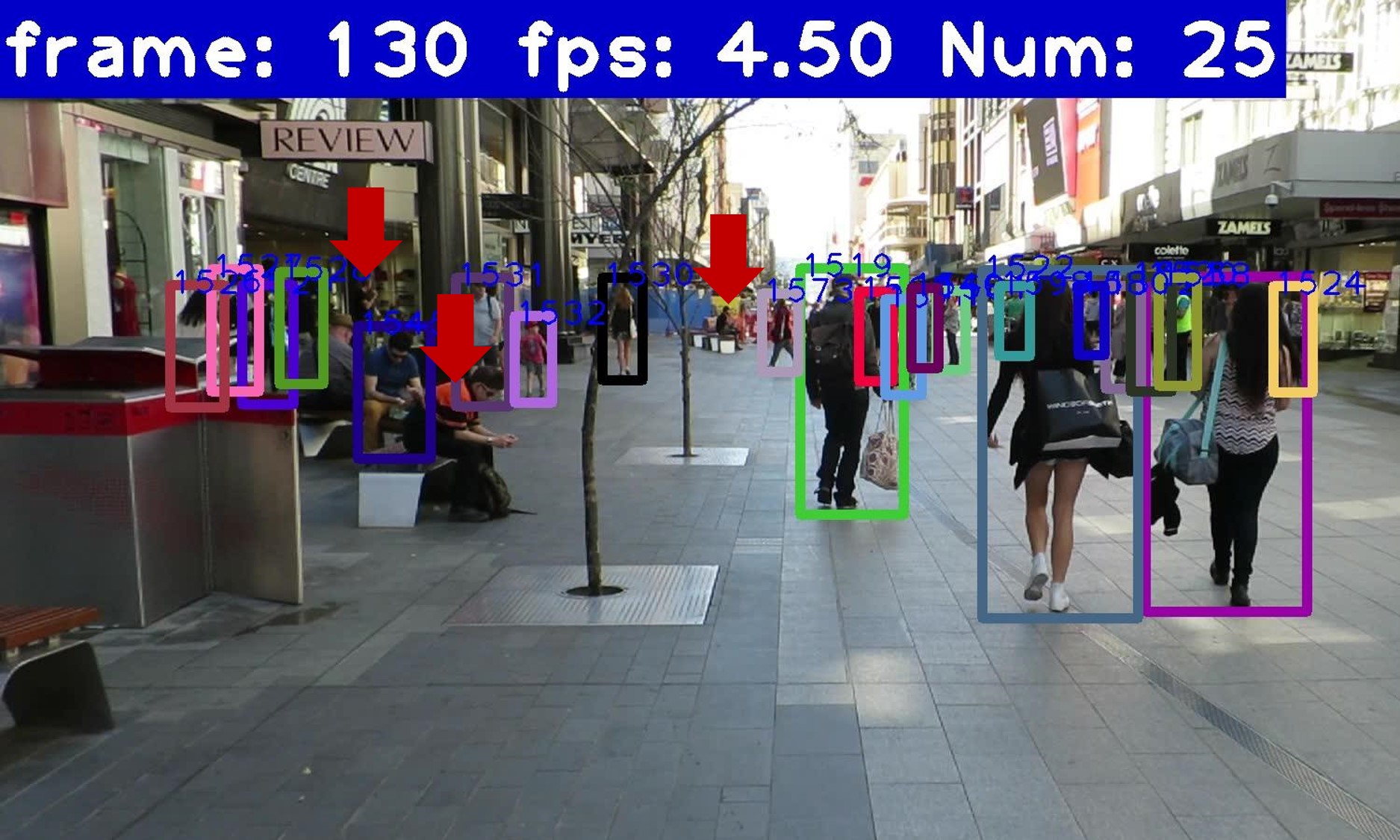}} \hspacefigure
\subfloat{\includegraphics[width=\widththird]{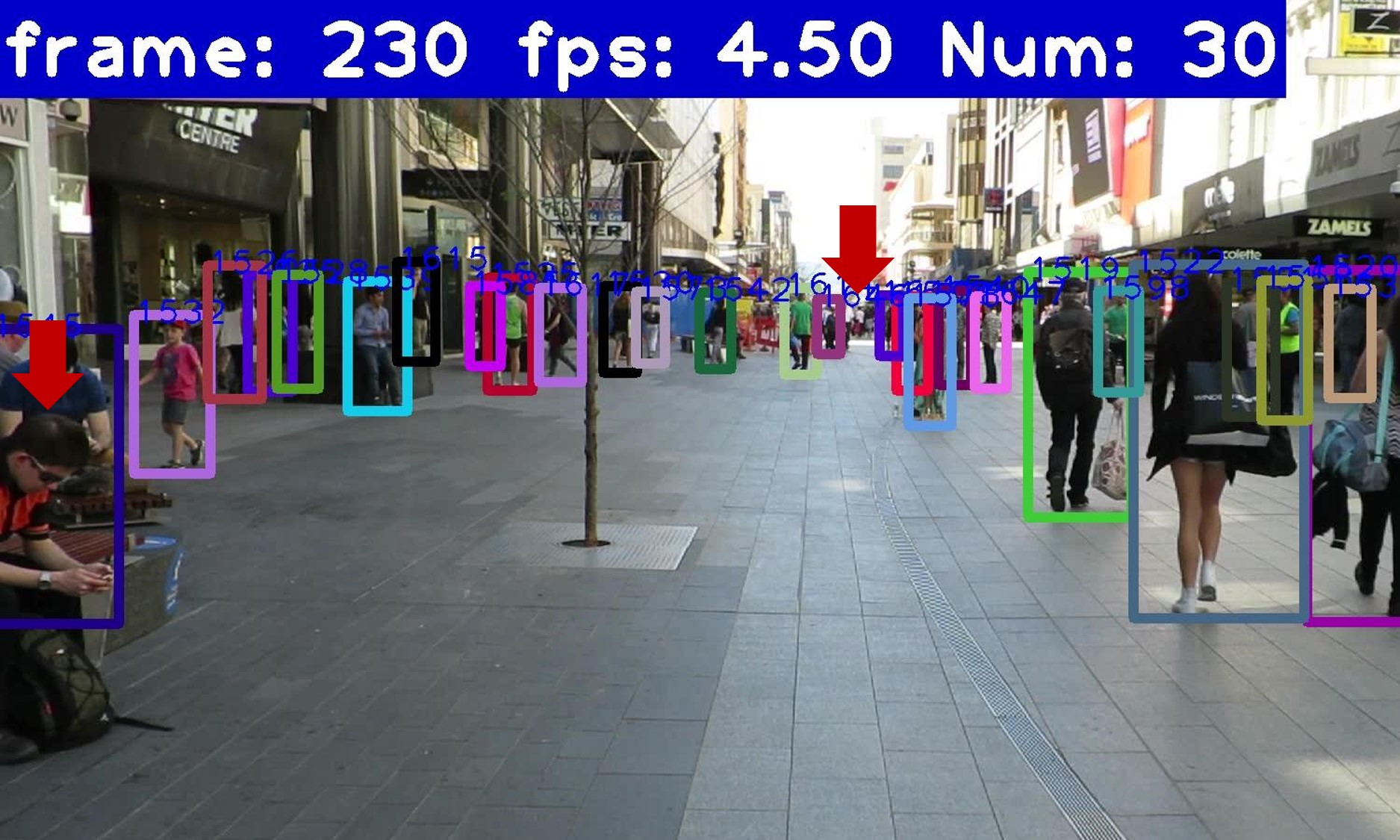}} \hspacefigure
\subfloat{\includegraphics[width=\widththird]{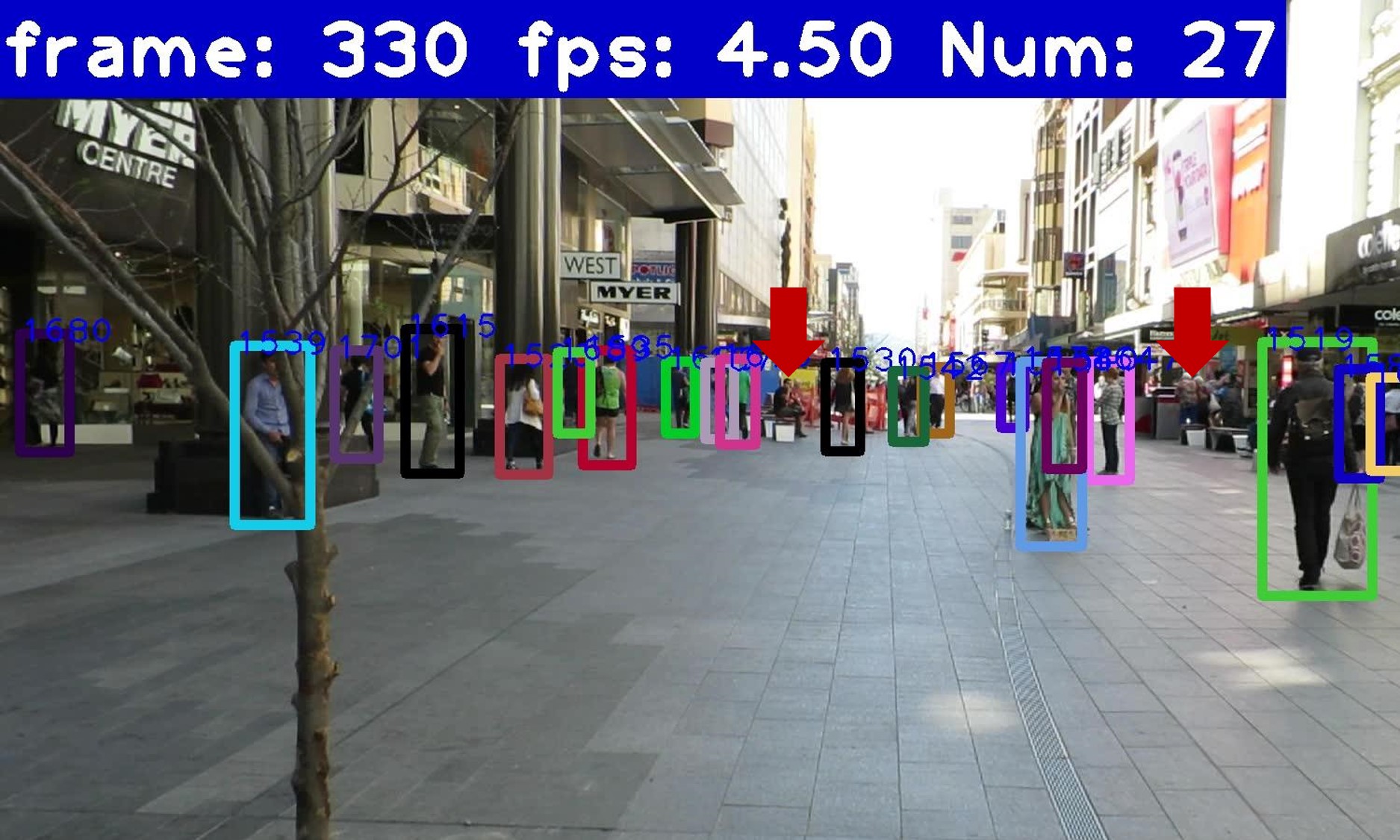}} \hspacefigure
{\small CSTrack~\cite{liang2022rethinking}~(TIP 2022)} \\

\vspace{-0.10in}
\subfloat{\includegraphics[width=\widththird]{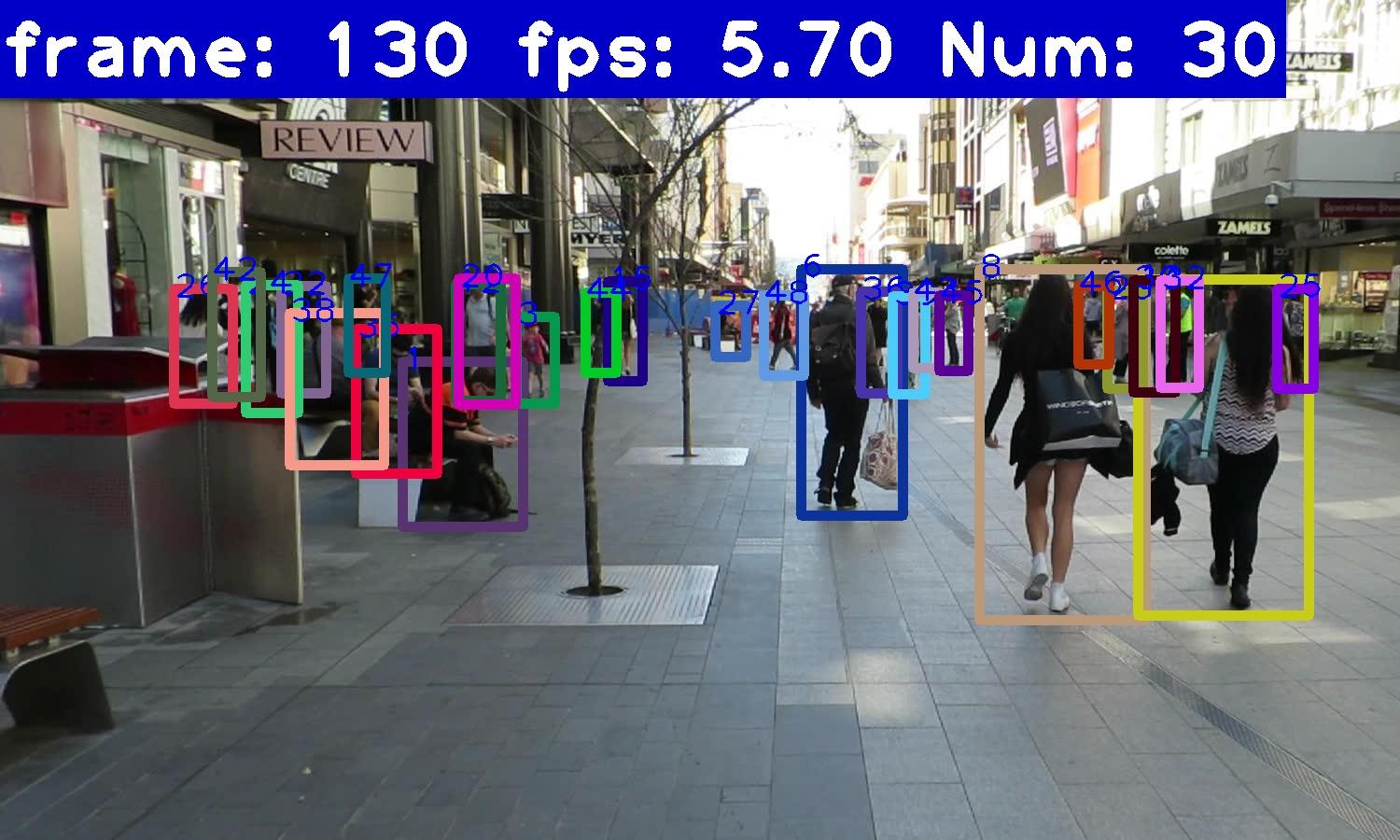}} \hspacefigure
\subfloat{\includegraphics[width=\widththird]{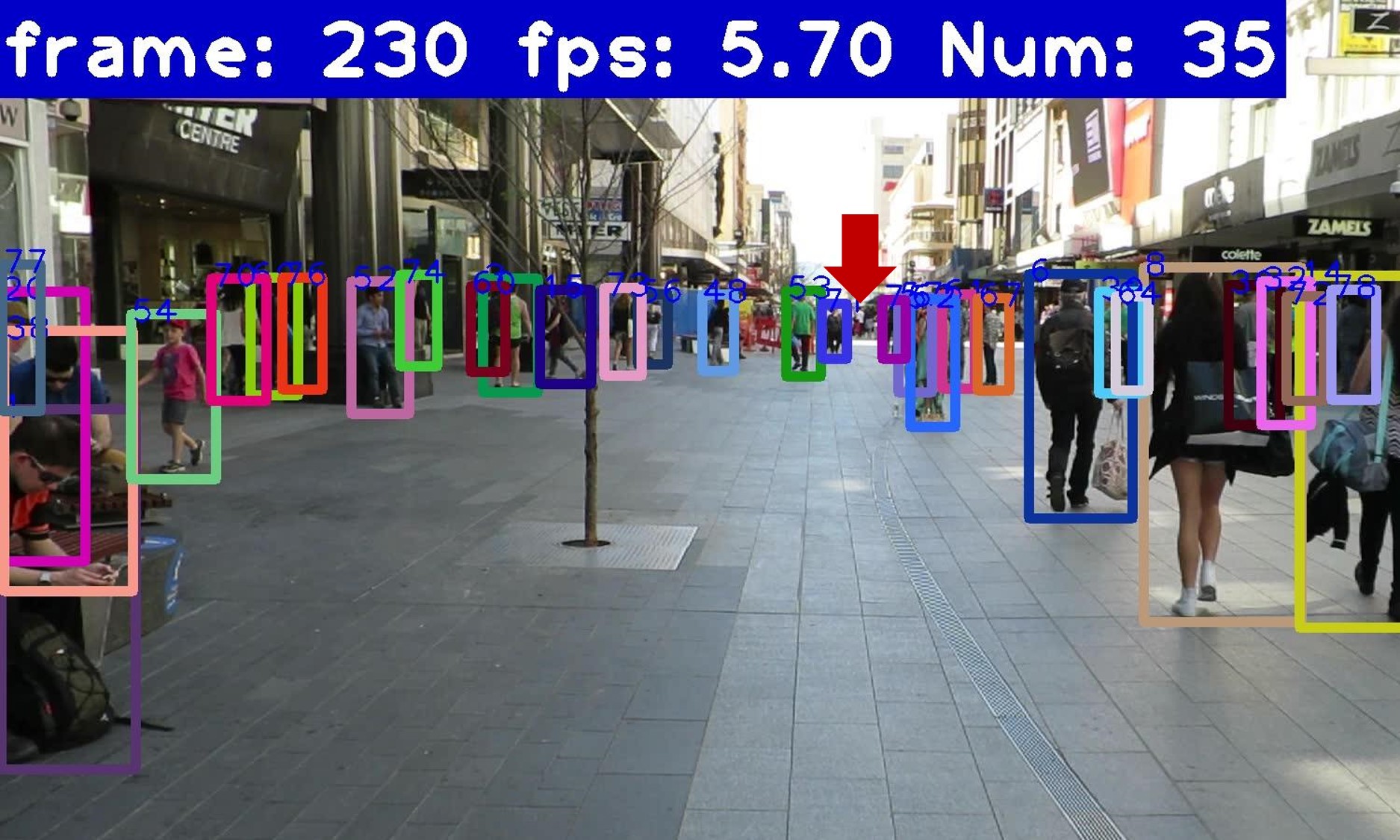}} \hspacefigure
\subfloat{\includegraphics[width=\widththird]{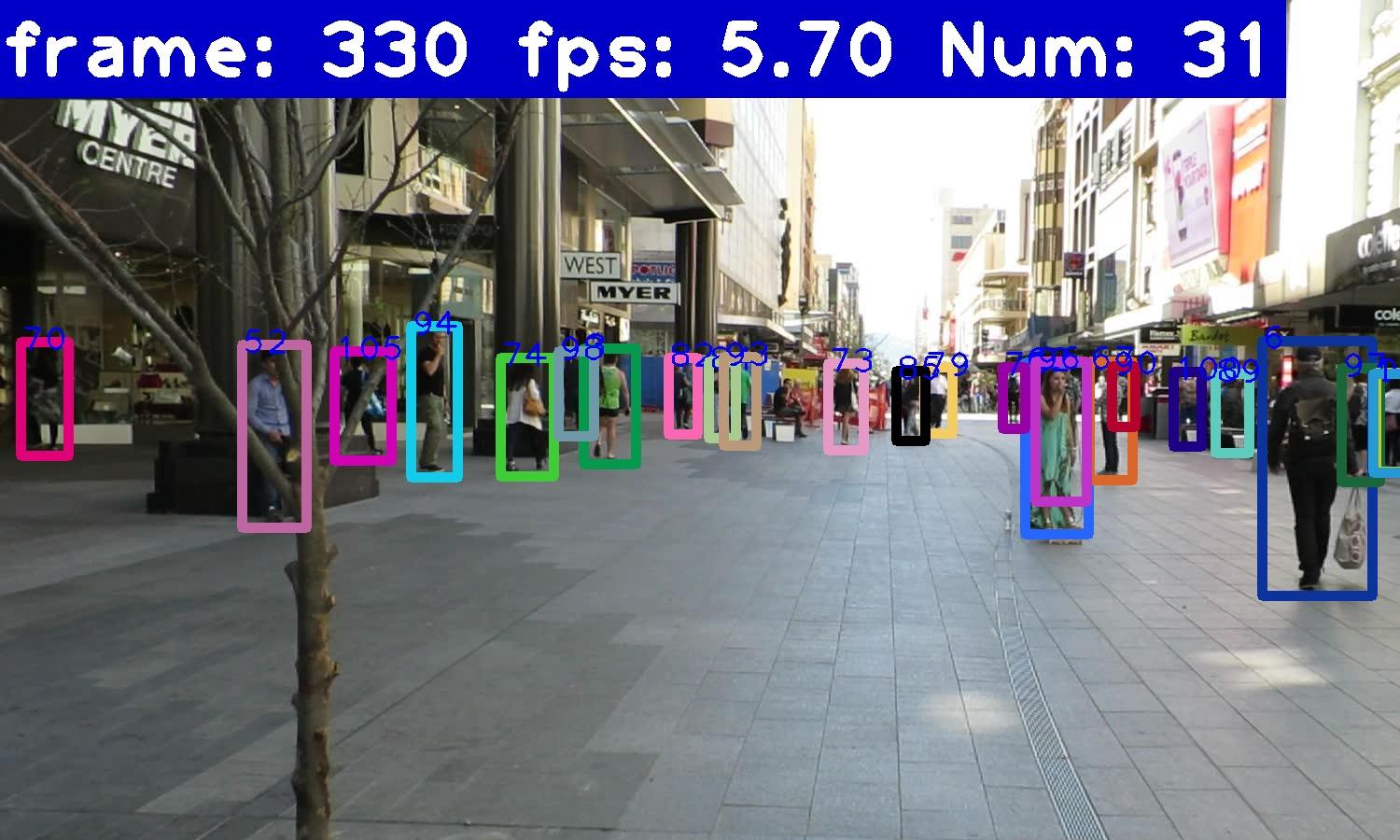}} \hspacefigure
{\small Trackformer~\cite{meinhardt2022trackformer}~(CVPR 2022)} \\

\vspace{-0.10in}
\subfloat{\includegraphics[width=\widththird]{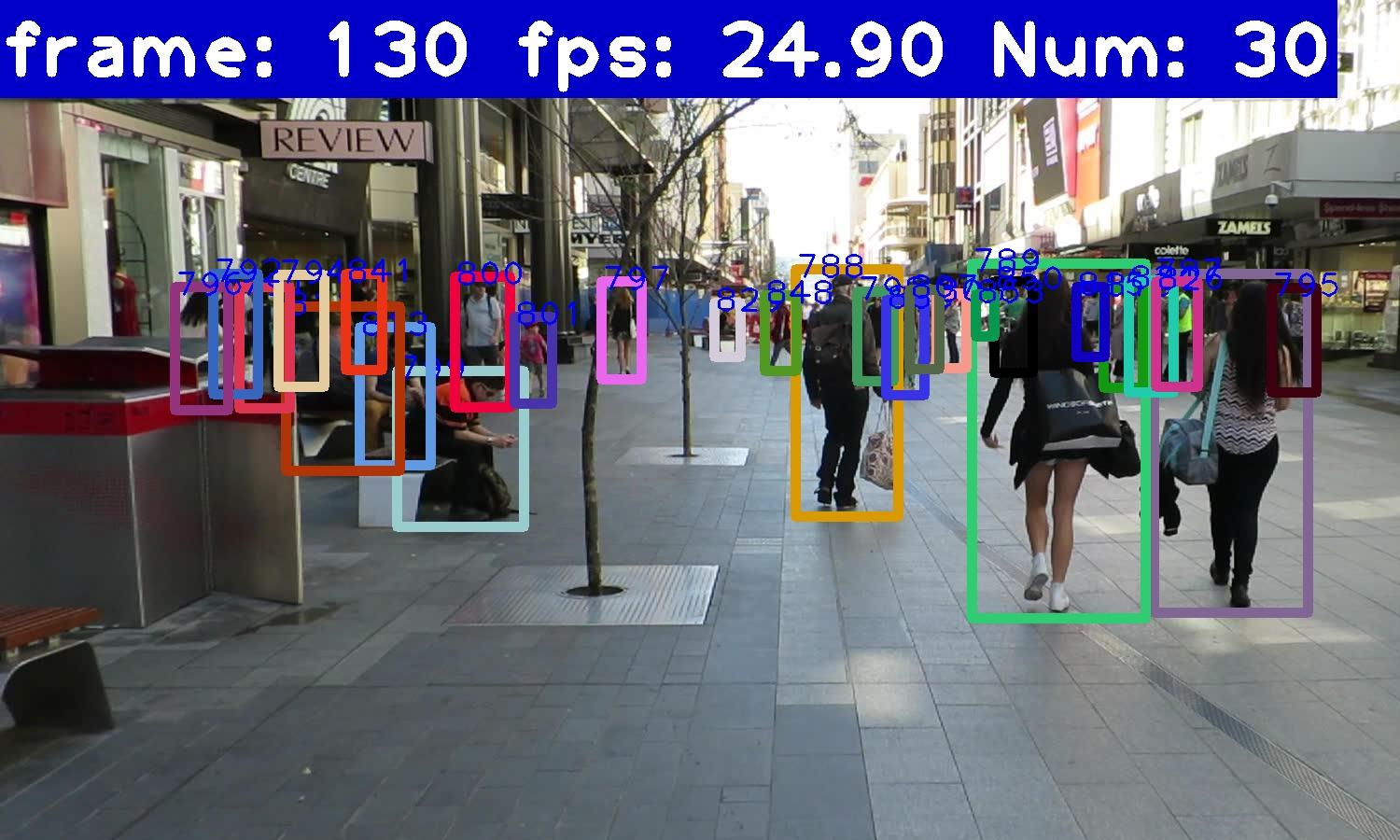}} \hspacefigure
\subfloat{\includegraphics[width=\widththird]{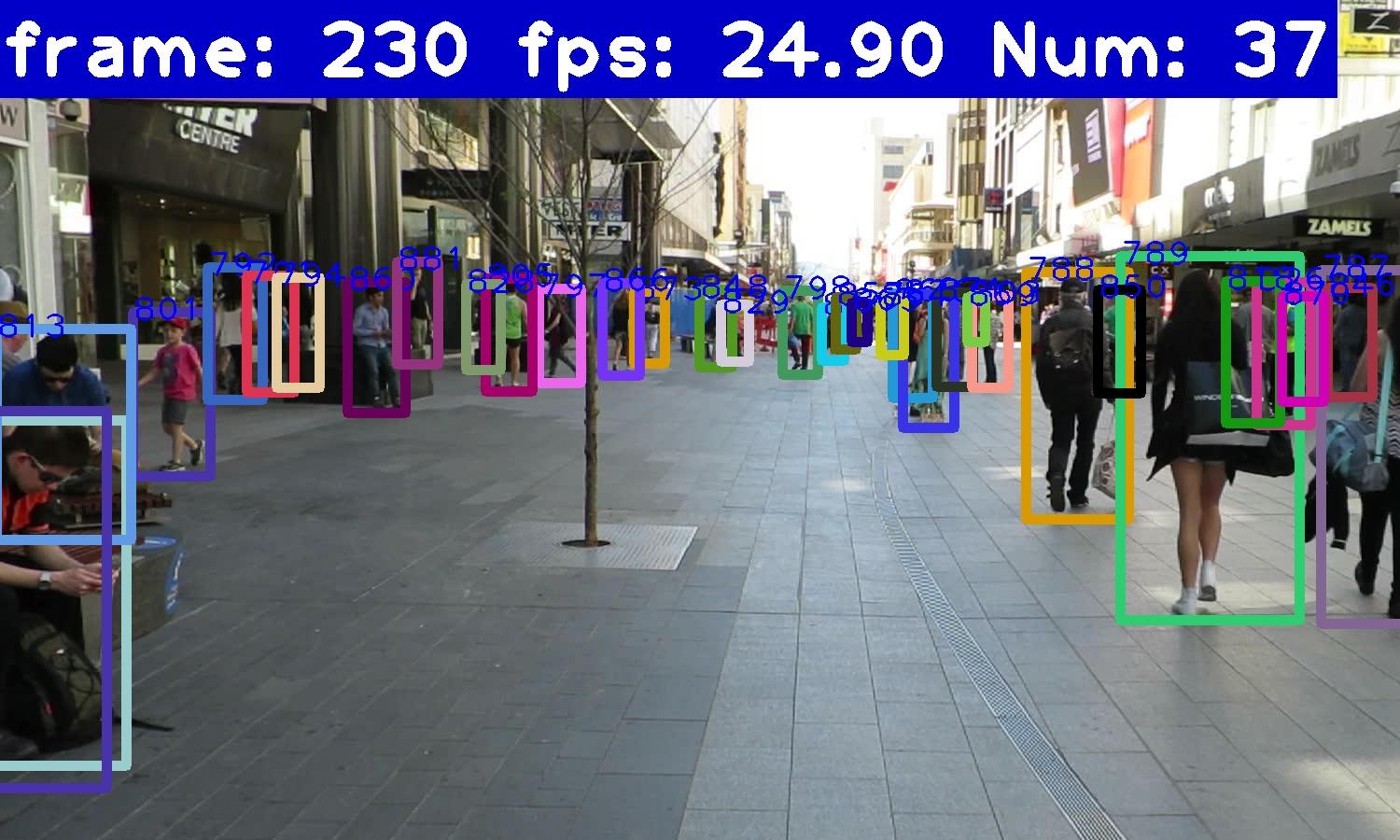}} \hspacefigure
\subfloat{\includegraphics[width=\widththird]{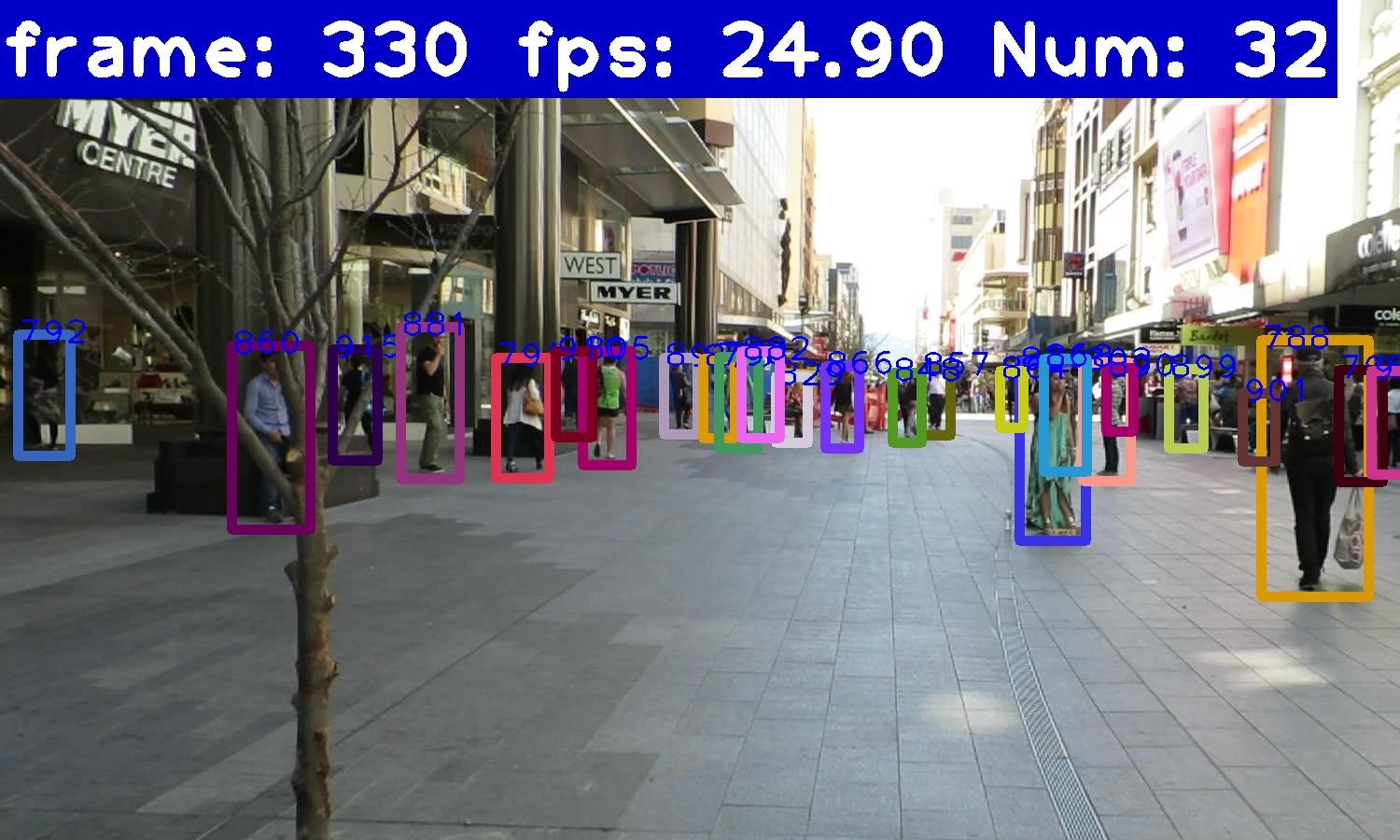}} \hspacefigure \\
{\small CountingMOT (ours)} \\

\caption{Qualitative results on \emph{MOT17-07} test set. 
As observed, all the trackers except ours loss object detections (marked with red arrows) and thus have relatively worse tracking performance. Also, our tracker has more object count.}\label{fig:ExpMOT17-07}
\vspace{-0.2in}
\end{figure*}

TBC~\cite{ren2020tracking} also adopts crowd density map to improve object detections, but it splits density map estimation and object detection into two separate steps. This means that object detections are highly \new{dependent} on the quality of density map, and they can't be optimized simultaneously, which affects the tracking performance (MOTA of 53.9) and running speed (FPS of 6.7) significantly. TBC needs to build a whole graph across all the frames, which makes it impossible for real applications. In Fig.~\ref{fig:ExpMOT17-07}, we also show some qualitative results on \emph{MOT17-07} test set. From left to right, frames are 130, 230 and 330, respectively. As observed, FairMOT~\cite{zhang2021fairmot} and CSTrack~\cite{liang2022rethinking}~lose object detections in some occluded situations (marked with red arrows) and thus have relatively worse tracking performance. Trackformer~\cite{meinhardt2022trackformer} can accurately locate occluded persons, but it may miss object detections in the far distance (zoom in for clear visualization). Using crowd density map as auxiliary information, our tracker has more object detections.

\begin{figure*}[!hbtp]
\centering
\captionsetup[subfigure]{labelformat=empty}

\vspace{-0.10in}
\subfloat{\includegraphics[width=\widththird]{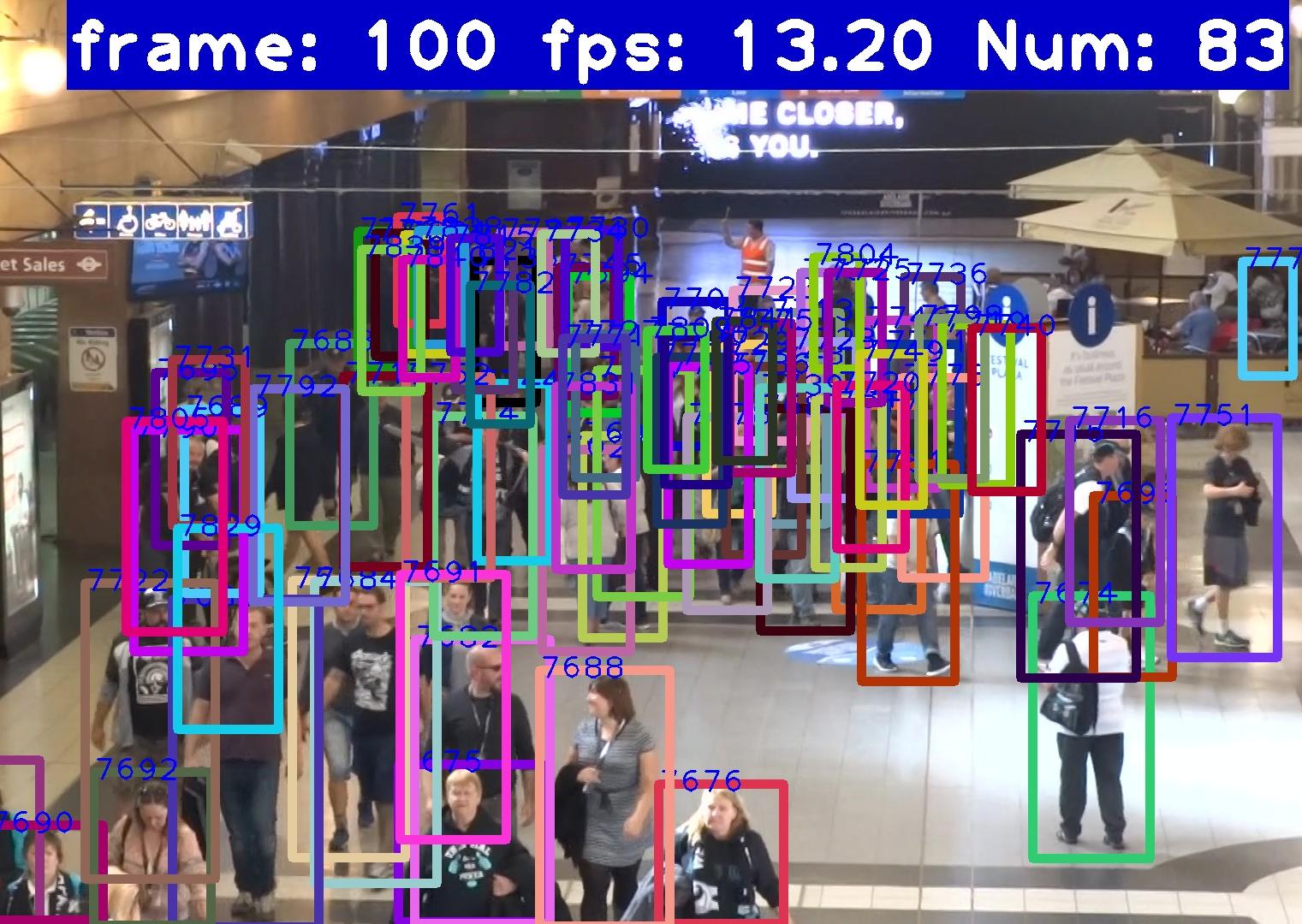}} \hspacefigure
\subfloat{\includegraphics[width=\widththird]{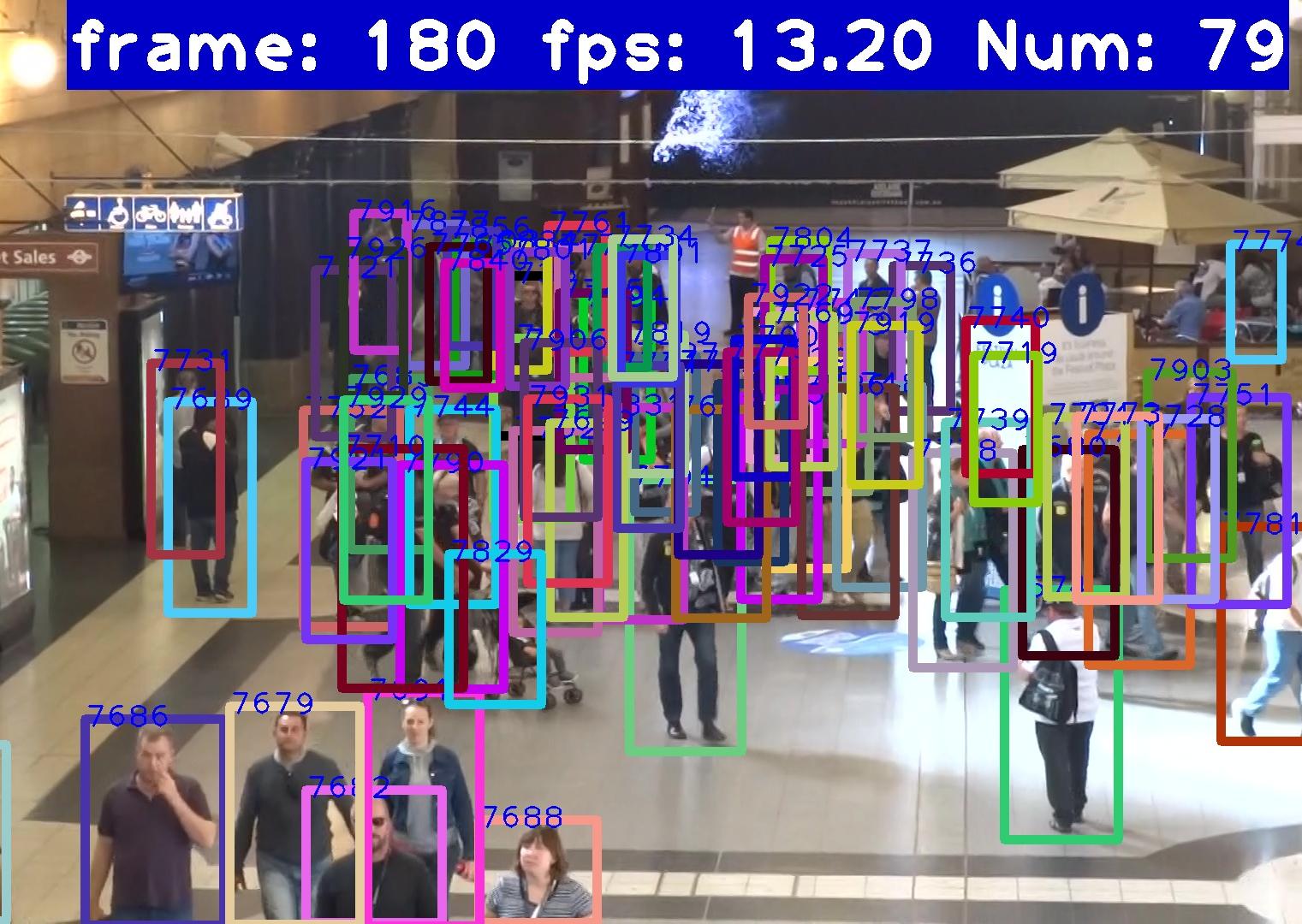}} \hspacefigure
\subfloat{\includegraphics[width=\widththird]{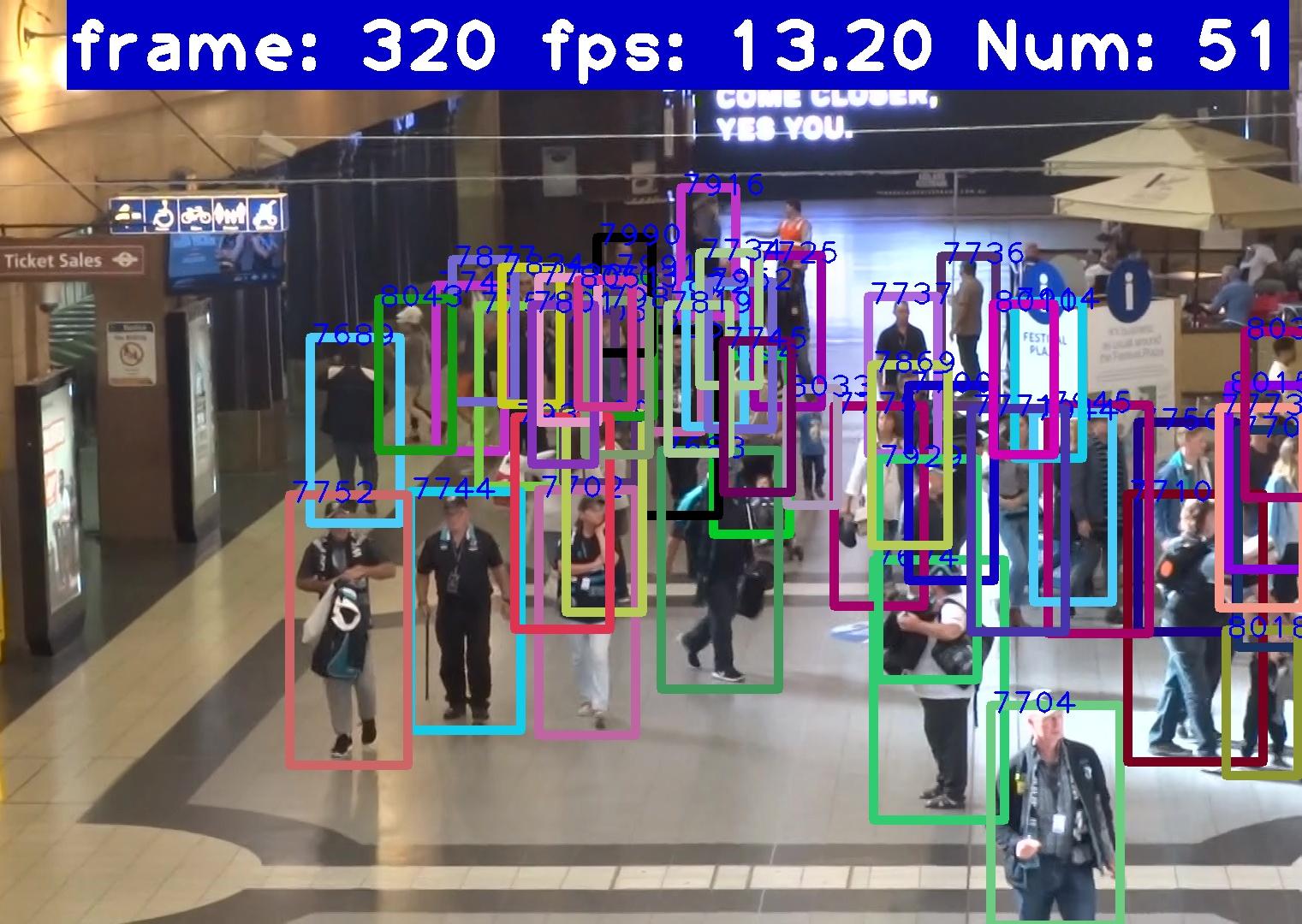}} \hspacefigure
{\small  FairMOT~\cite{zhang2021fairmot}~(IJCV 2021)} \\

\vspace{-0.10in}
\subfloat{\includegraphics[width=\widththird]{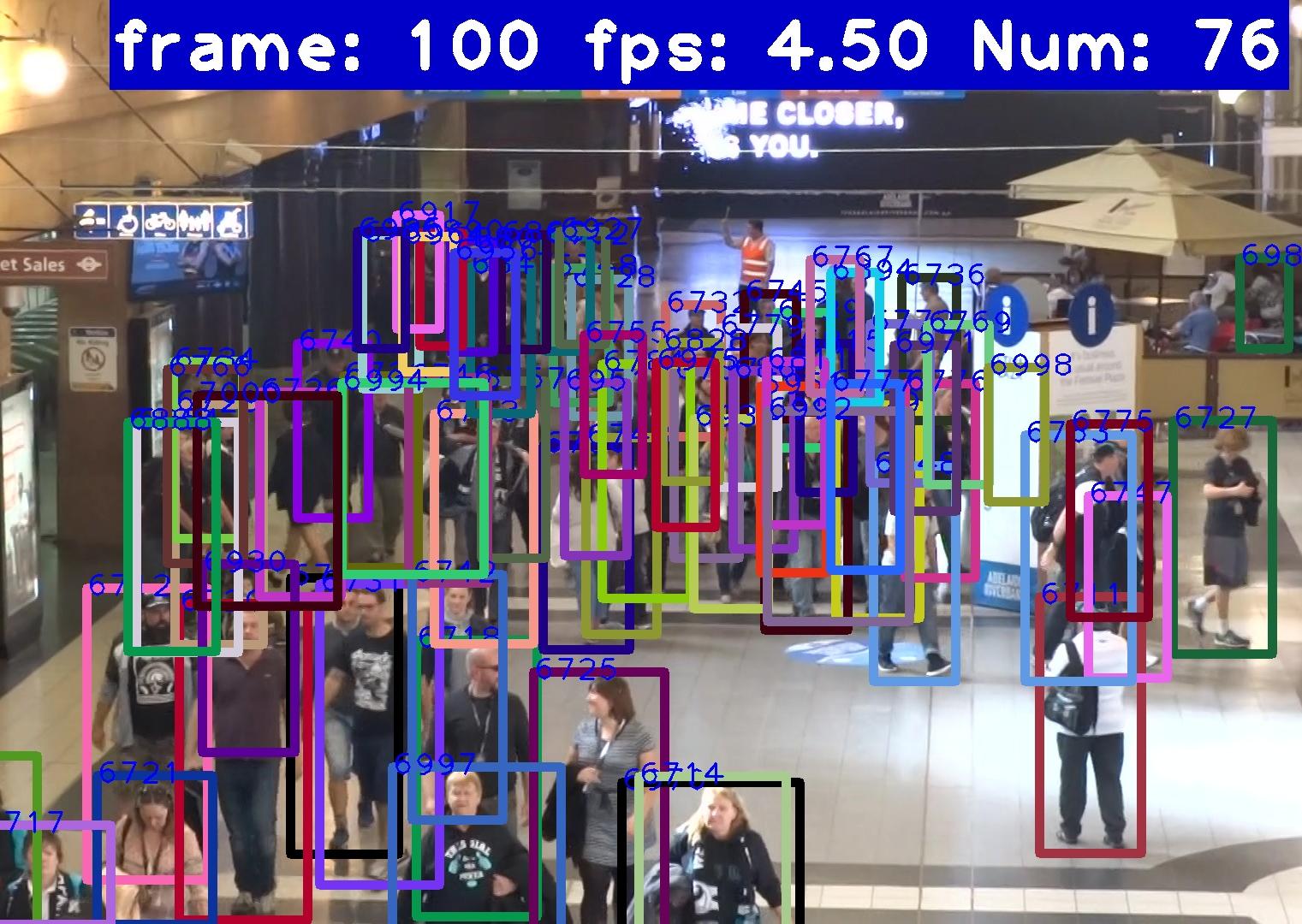}} \hspacefigure
\subfloat{\includegraphics[width=\widththird]{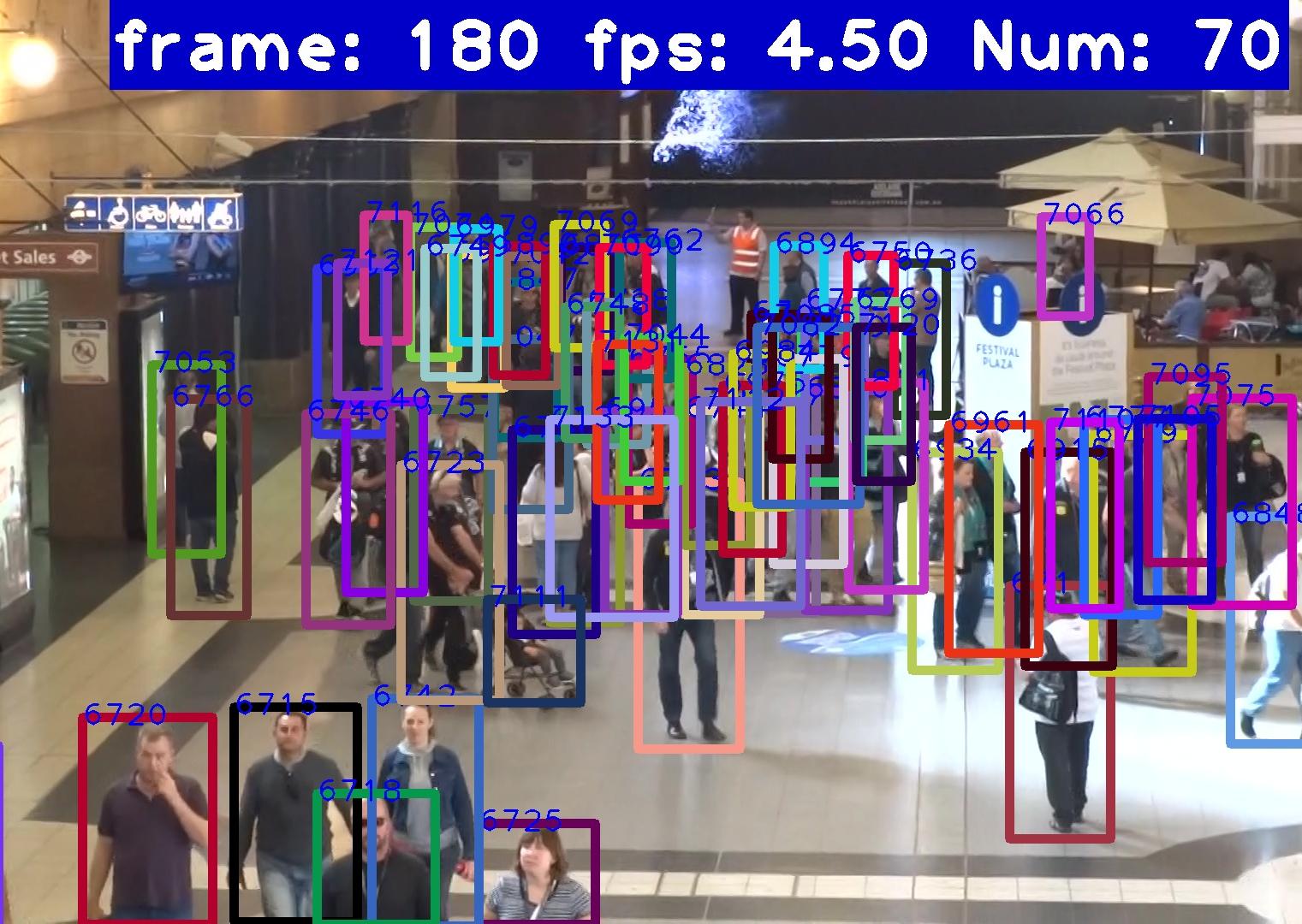}} \hspacefigure
\subfloat{\includegraphics[width=\widththird]{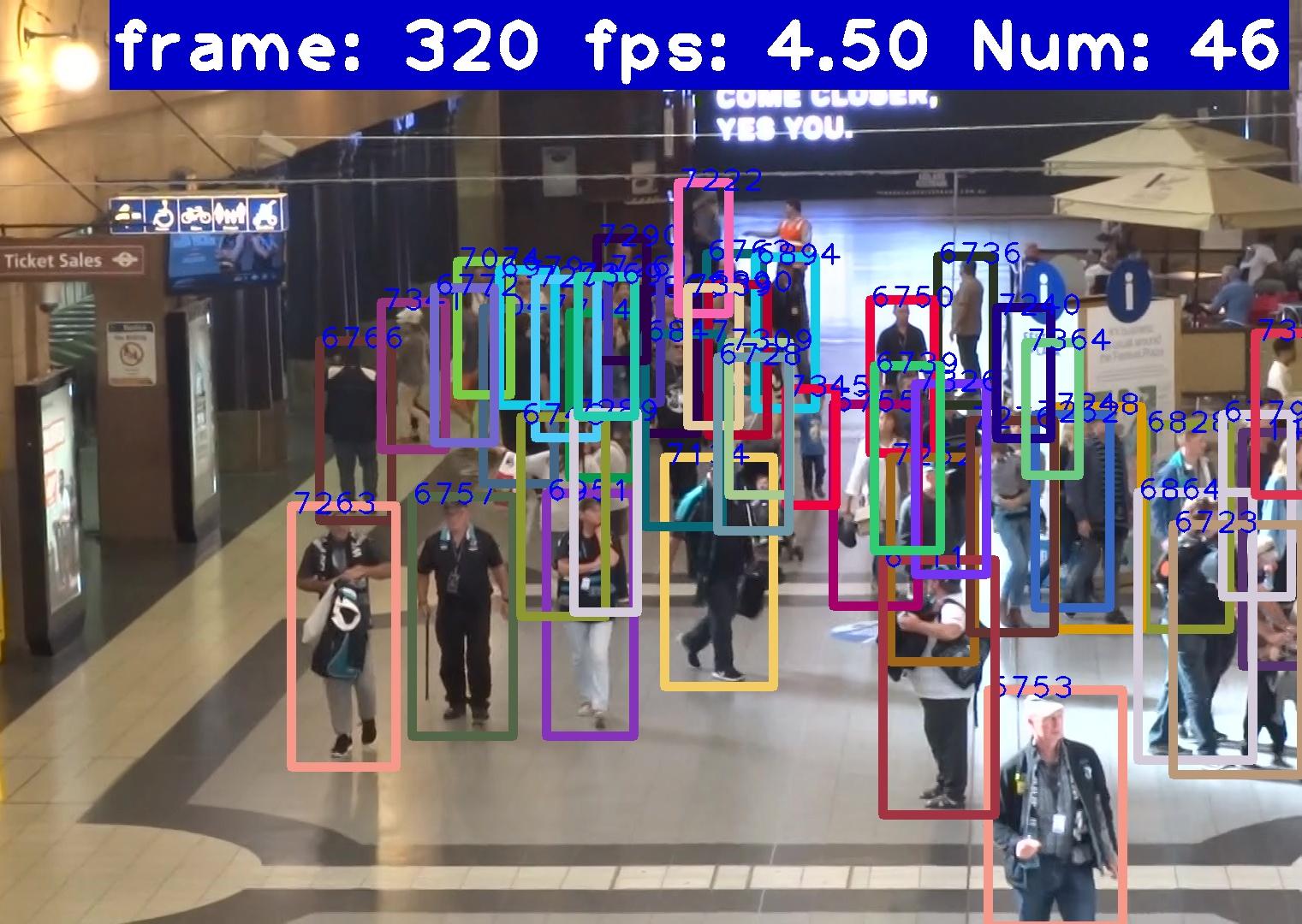}} \hspacefigure
{\small CSTrack~\cite{liang2022rethinking}~(TIP 2022)} \\

\vspace{-0.10in}
\subfloat{\includegraphics[width=\widththird]{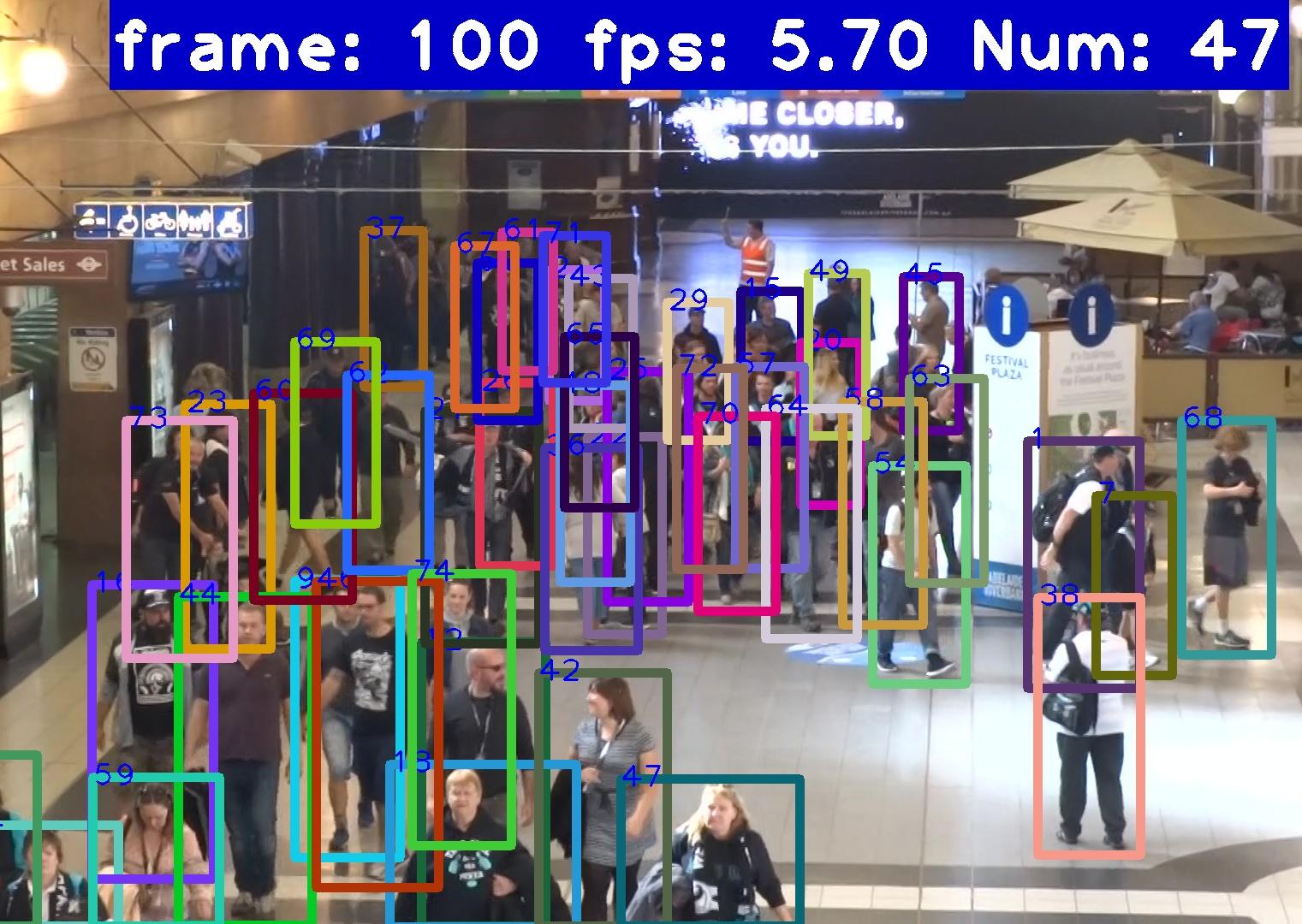}} \hspacefigure
\subfloat{\includegraphics[width=\widththird]{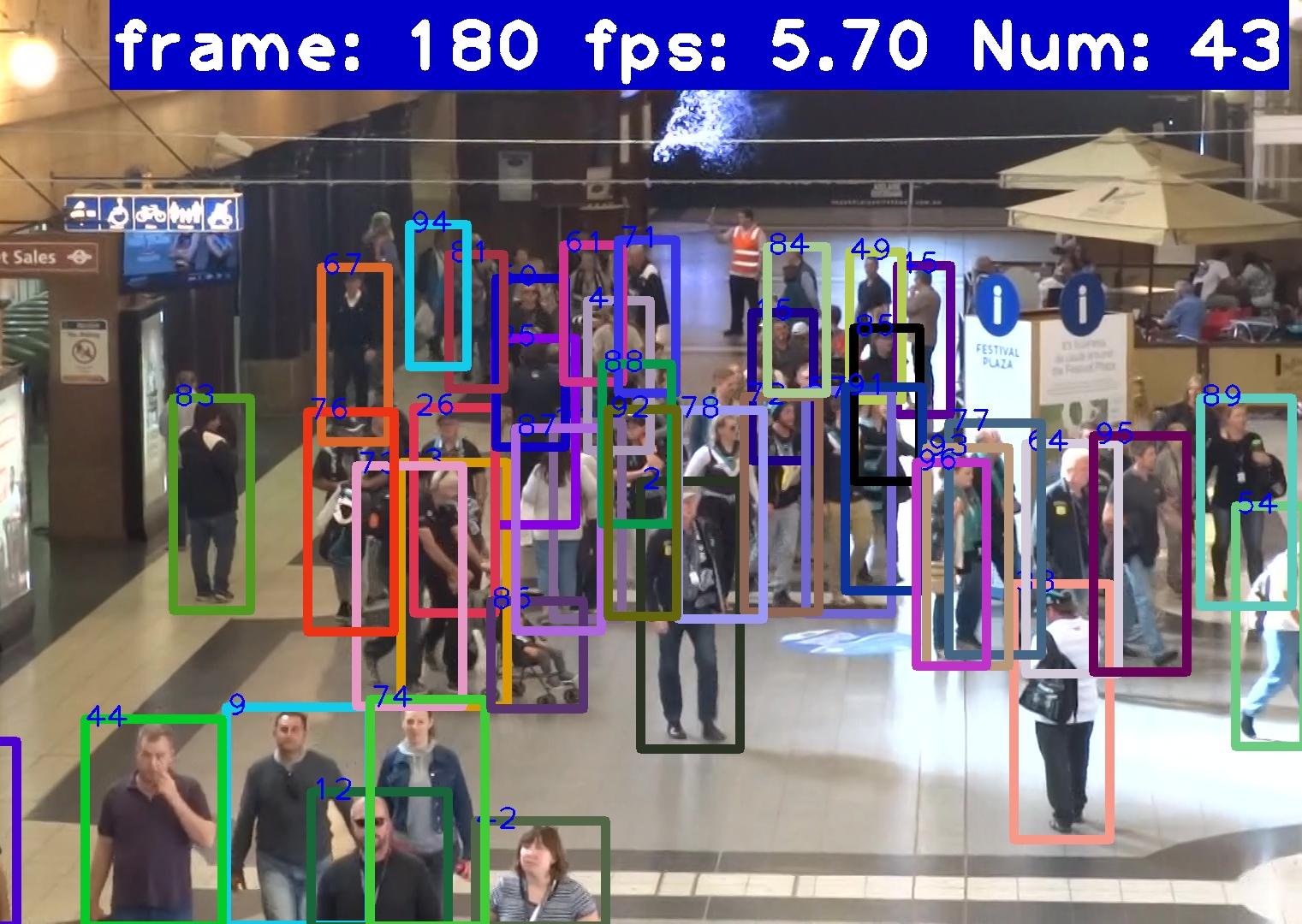}} \hspacefigure
\subfloat{\includegraphics[width=\widththird]{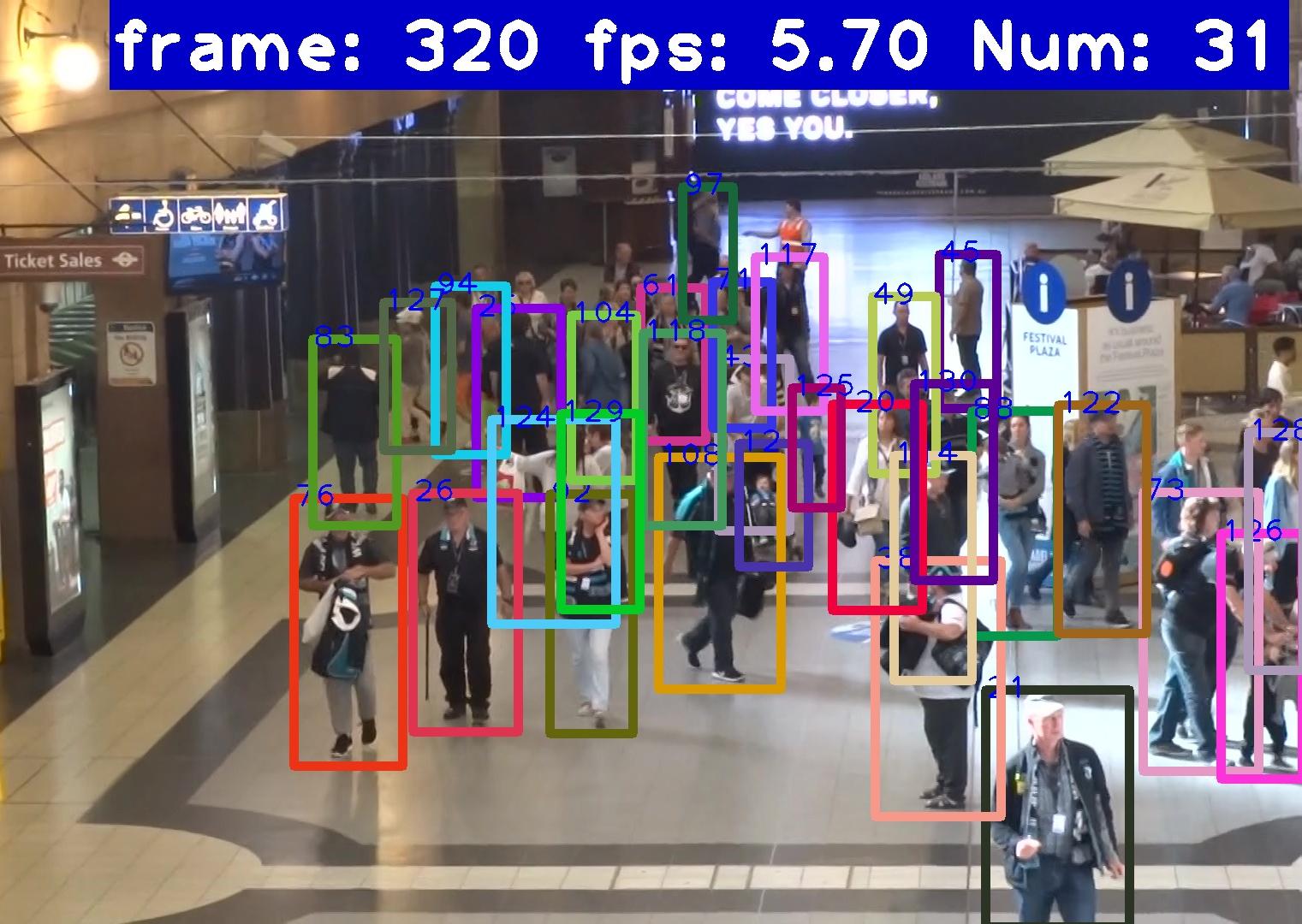}} \hspacefigure
{\small Trackformer~\cite{meinhardt2022trackformer}~(CVPR 2022)} \\

\vspace{-0.10in}
\subfloat{\includegraphics[width=\widththird]{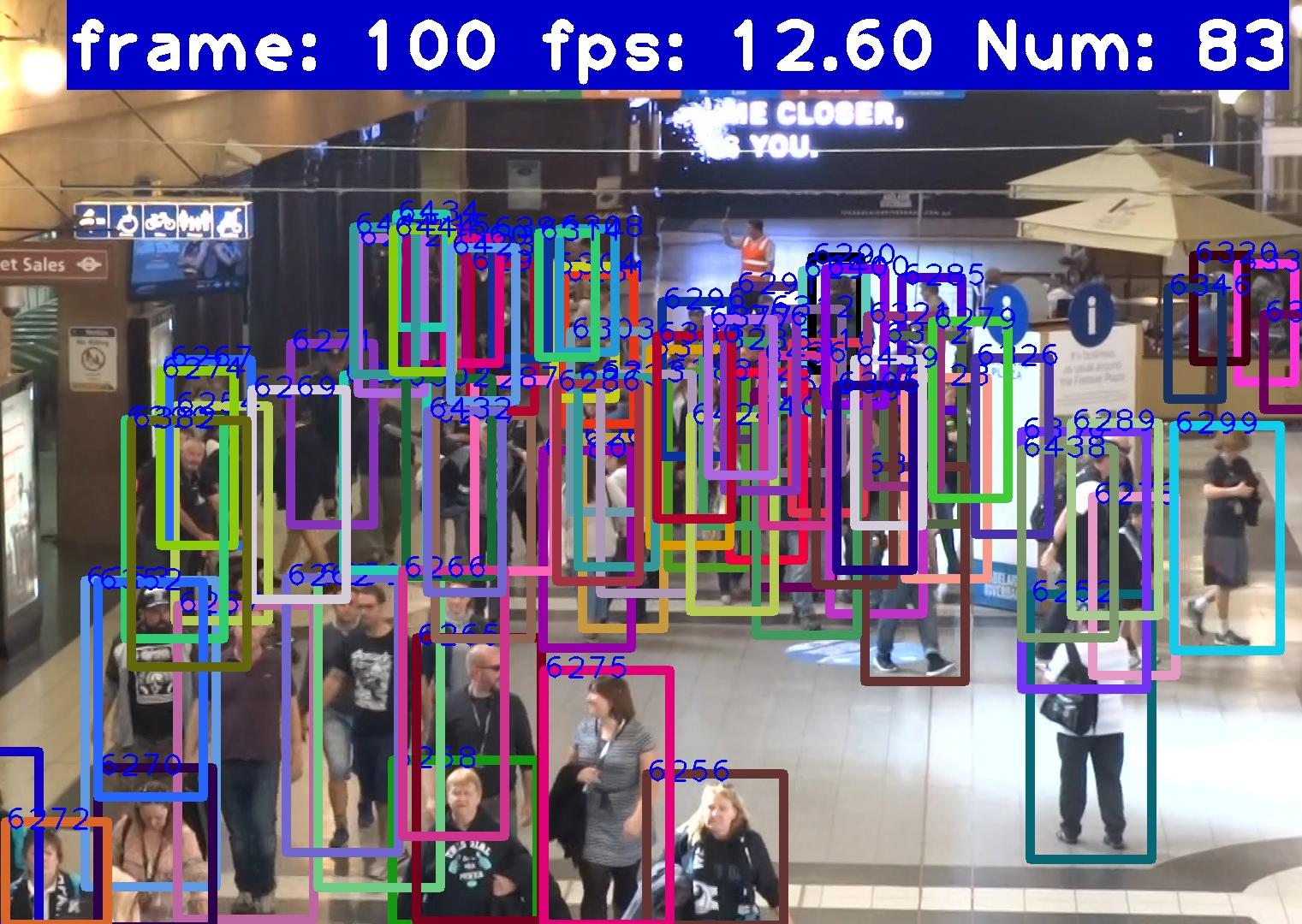}} \hspacefigure
\subfloat{\includegraphics[width=\widththird]{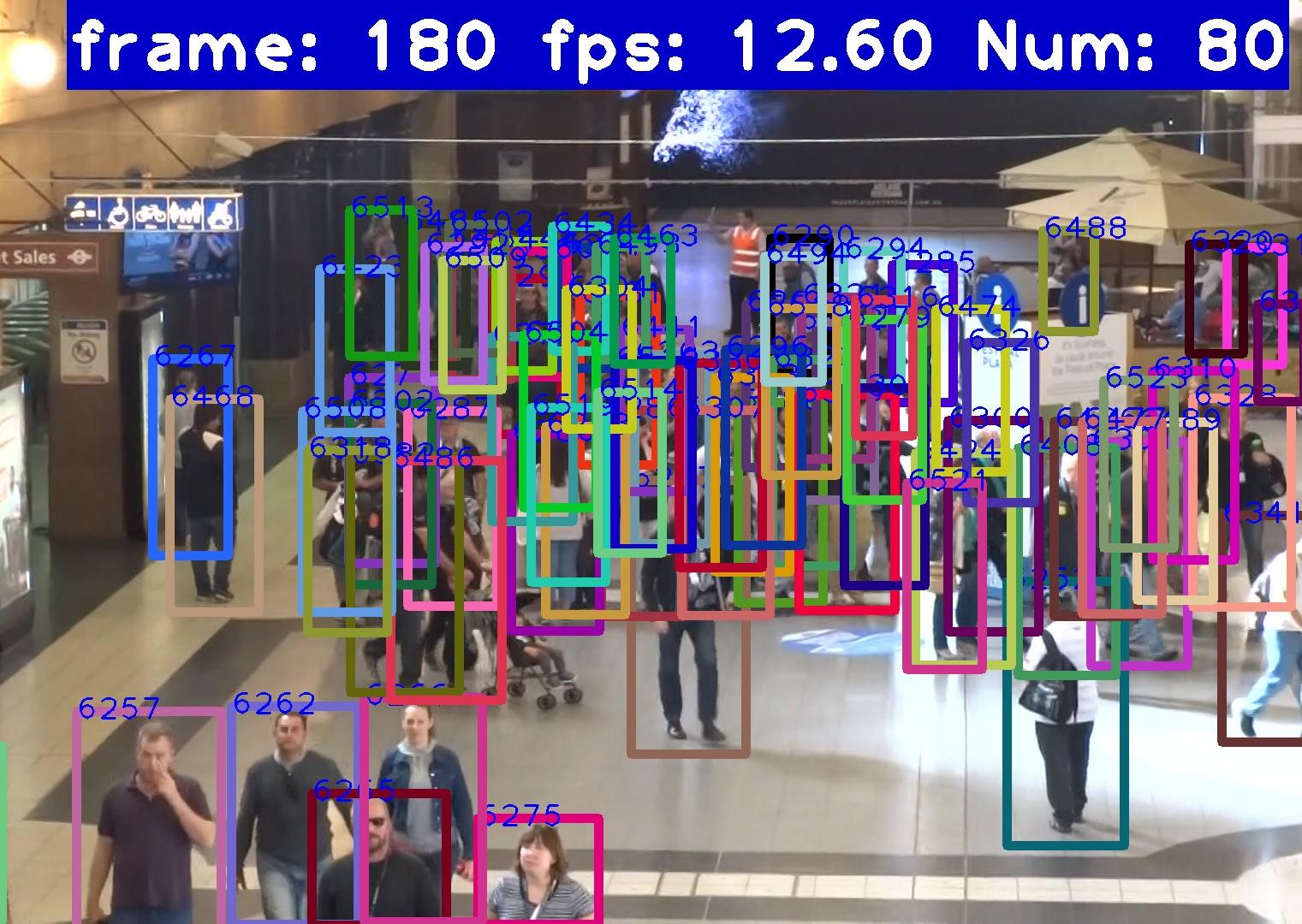}} \hspacefigure
\subfloat{\includegraphics[width=\widththird]{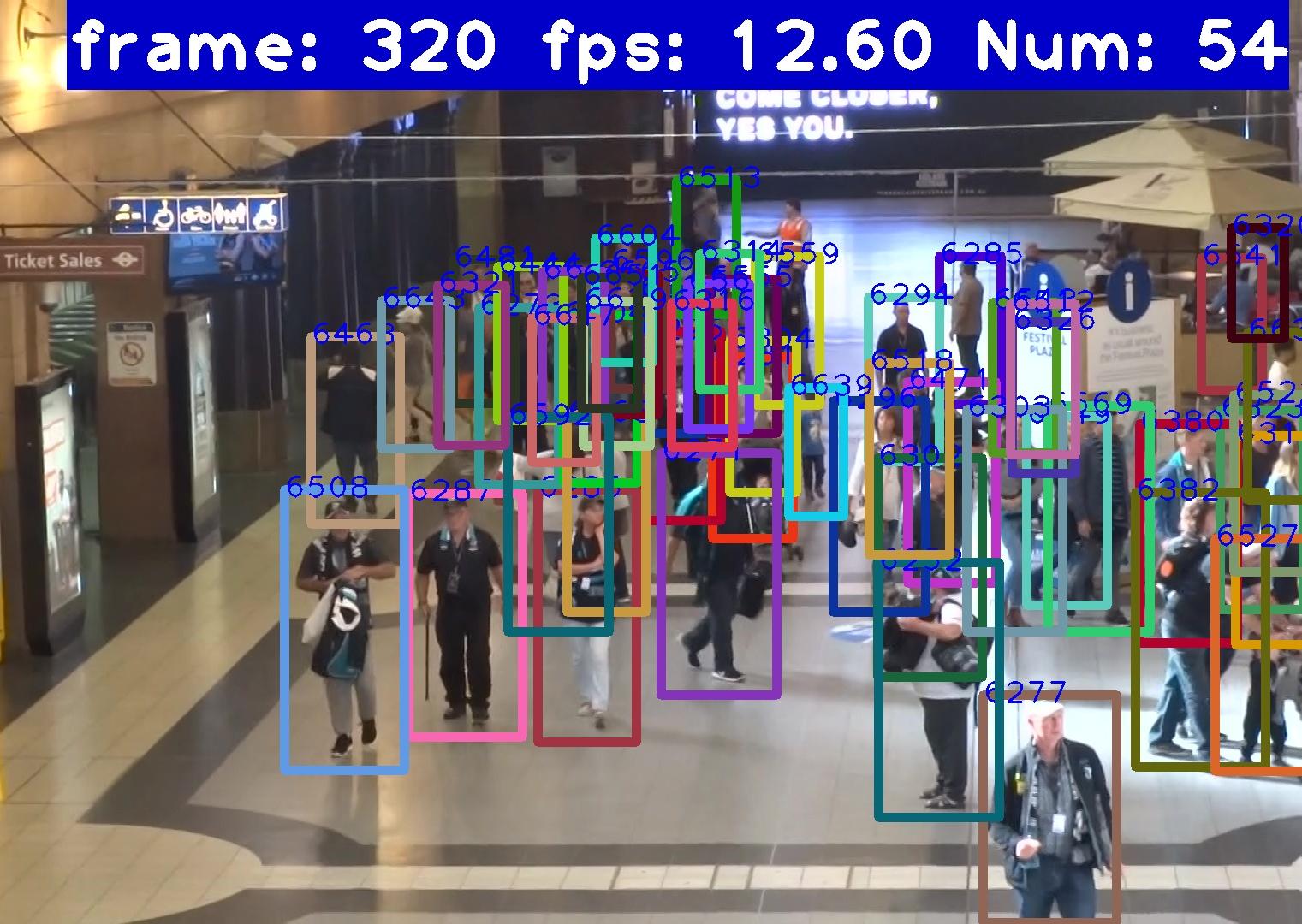}} \hspacefigure \\
{\small CountingMOT (ours)} \\

\caption{Qualitative results on \emph{MOT17-20} test set. For the crowd scene, CSTrack~\cite{liang2022rethinking} and Trackformer~\cite{meinhardt2022trackformer} lose too many object detections (see the ``Num'' in the figure), while our CountingMOT tracker has the most object count, which implicitly indicates that crowd density map indeed helps to locate occluded persons (zoom in for clear visualization).}\label{fig:ExpMOT20-07}
\vspace{-0.2in}
\end{figure*}

For the extremely crowd scenes \emph{MOT20}, our method can further show its superior. E.g., CountingMOT achieves higher MOTA (70.2) and IDF1 (72.4). In our model, the density map can help the detection task to find occluded or missed detections. In return, the detection task can also enhance the localization ability of crowd density map. The two tasks can jointly improve the detection results, 
and thus can improve the tracking results. Compared with FairMOT~\cite{zhang2021fairmot}, IDF1 is increased while IDs is significantly decreased (IDs of 2,795 vs 5,243).  \new{
It is interesting that the FP of CountingMOT is significantly decreased from $103,440$ to $33,531$, but the FN is increased from $88,901$ to $117,886$. Crowd density map focuses on counting of the whole scene, and it can help to locate objects well in sparse scenes. However, for crowd scenes, the quality of the crowd density map will be affected, and thus the localization ability is weakened. 
For crowd scenes in \emph{MOT20}, FairMOT introduces many false detections (FP of 103,440), and the estimated crowd density map can reject false detections using counting constraint~(FP of 33,531 for CountingMOT). However, crowd density map also causes an increase of FN. Overall, the detection result is improved (MOTA of 70.2 vs 61.8). 
Also, the CountingMOT performs better than the recent two trackers MeMOT~[69] and  Trackformer~[71]~(e.g., MOTA of 70.2 vs 66.1 and 66.8).} In Fig.~\ref{fig:ExpMOT20-07},  we also show some qualitative results on \emph{MOT17-20} test set. CSTrack~\cite{liang2022rethinking} and Trackformer~\cite{meinhardt2022trackformer} focus more on data association, and they work well in sparse scenes (see Fig.~\ref{fig:ExpMOT17-07}). However, for the crowd scene, they lose too many object detections (see the ``Num'' in the figure). Our CountingMOT tracker has the most object count, which implicitly indicates that crowd density map indeed helps to locate occluded persons. 
Please see details in MOT challenge (our tracker is denoted as ``CountingMOT''). 

\new{
Through the analysis of Tab.~\ref{table:sota}, our CountingMOT model tries to find a balance between the object detection task and the crowd density map estimation task. For sparse scenes, the crowd density map has a relatively strong localization ability, and it can help object detection task to decrease FN. For crowd scenes, the localization ability of crowd density map is weakened, but it can also help object detection task to reject false detections. Overall, the joint detection and counting indeed help to improve MOT performance (from MOTA and IDF1).}

\rwh{Recently, some methods use the high-performance detector YOLOX~\cite{ge2021yolox} to build MOT trackers, and have achieved significant progress~\cite{qin2023motiontrack,cetintas2023unifying,zhang2022bytetrack}. To further validate our CountingMOT model, we change the backbone DLA-34~\cite{zhou2020tracking} to YOLOX~\cite{ge2021yolox} (denoted as ``CountingMOT\_v2"), which adopts Feature Pyramid Network (FPN) to extract multi-level features (denoted as ``P3", ``P4" and ``P5") for object detection. Considering the feature resolution, we choose the ``P4" level (1/16 of the input image) to build counting and reID tasks. The training details remains the same with the original YOLOX. The tracking results are reported in the gray part of Tab.~\ref{table:sota}. For \emph{MOT17}, CountingMOT\_v2 performs better than other YOLOX based trackers  Motion\_Track~\cite{qin2023motiontrack}, SUSHI~\cite{cetintas2023unifying} and ByteTrack~\cite{zhang2022bytetrack} in terms of MOTA. However, it achieves relatively worse results than UTM~\cite{you2023utm} which proposes a identity-aware module to improve long-term associations. UTM also reports a very high FPS, and the reason may be it only calculate the association time. For \emph{MOT20}, CountingMOT\_v2 achieves the best MOTA, which indicates that the joint modelling of counting, detection and reID is effective for MOT.}

\subsection{Effect of sliding windows size $\textbf{w}_{k}$} \label{sec:IV-D}
In (\ref{eq8}), the sliding window size $\textbf{w}_k$ determines the region to count the objects, which will affect the final loss ${{\mathcal{L}}_{\text{cnt-cd}}}$. In other words, the size of the sliding window size $\textbf{w}_k$ means how much prior detection information is used for crowd density estimation. A large sliding window contains more object counts, but it weakens the constraint of localization for crowd density map. By contrast, a small sliding window can improve the location ability, but it ignores the counting ability of crowd density map. For the experiment, we train all the
models on half of the training set and validate on another
half (demoted as ``validation set''). In Tab.~\ref{table:window}, we summarize the results on the validation set of \emph{MOT17} by using different sliding window size $\textbf{w}_k$. As observed, the MOTA continuously increases when the sliding window size changes from $7$ to $19$. However, the MOTA drops when the sliding window size becomes larger than $19$. For a small sliding window, it focuses more on individual detection and thus ignores the overall understanding of the scene. A large sliding window introduces more detection prior but the counting result is degraded. According to the ablation study, we choose $\textbf{w}_k=19$ in our work.

\begin{table}
\begin{center}
\setlength{\tabcolsep}{5pt}
\caption{Comparison of using different sliding window size $w_k$ on the validation set of \emph{MOT17}. The best results are shown in {\bf bold}.}
\label{table:window}
\begin{tabular}{lcccccc}
\toprule
Sliding window size & MOTA$\uparrow$ & IDF1$\uparrow$ & MT$\uparrow$& ML$\downarrow$ & IDS $\downarrow$\\
\midrule
$w_k = 7$ & 63.8 & 65.2 & 40.30\% &24.8\% & 432\\
$w_k = 9$ & 64.7 & 69.3 &43.63\%& 17.27\% & 406\\
$w_k = 11$ & 67.8 & {70.3} &45.15\%&16.97\%& 303&\\
$w_k = 13$ & 69.0 & 68.2 &45.45\%&16.97\%& 328\\
$w_k = 15$ & 68.3 & {70.2} & 44.85\% & 16.67\%&312\\
$w_k = 17$ &{69.5} & 75.6 &46.06\% & 16.97\% &{\bf 280}\\

{$w_k = 19$} & {\bf 71.8} & {\bf 76.3}&{\bf 49.55\%} &{\bf 13.27}\% & 309\\

$w_k = 21$ & 70.8 & 74.8 & {45.76\%} & {15.15\%}&345\\
$w_k = 23$ &68.8 & {74.3} & 44.85\% & 17.23\%& 361\\
\bottomrule
\end{tabular}
\end{center}
\vspace{-0.1in}
\end{table}

\subsection{Effect of amplification factor $\mu$}
 In (\ref{eq6}), the amplification factor $\mu$ affects the performance of crowd density map. A small $\mu$ leads to the non-convergence of the counting task, while a large $\mu$ affects the accuracy of the detection result. In Tab.~\ref{table:amplication}, we perform an ablation study of the amplification factor $\mu$ on the validation set. As observed, when $\mu =1$, the CountingMOT model achieves relatively worse tracking results (MOTA of 62.7). We check the predicted crowd density map, and find that it is a map of all zeros. This indicates that the counting task is hardly trained when $\mu$ is small. As $\mu$ increases, the tracking performance is continuously improved (e.g., MOTA is improved from 62.7 to 71.8). However, MOTA drops when $\mu$ becomes too large. The reason is that a large $\mu$ highlights the importance of counting while undermines object detection. Thus, we choose $\mu = 1000$ in our work, which is a good balance between counting and detection.
 \begin{table}
\begin{center}
\setlength{\tabcolsep}{5pt}
\caption{Effect of the amplification factor $\mu$ on the validation set of \emph{MOT17}. The best results are shown in {\bf bold}.}
\label{table:amplication}
\begin{tabular}{lcccccc}
\toprule
Amplification factor & MOTA$\uparrow$ & IDF1$\uparrow$ & MT$\uparrow$& ML$\downarrow$ & IDS $\downarrow$\\
\midrule
$\mu = 1$ & 62.7 & 65.2 & 40.30\% &24.8\% & 438\\
$\mu = 10$ & 66.5 & 69.4 &38.94\%& 17.11\% & 451\\
$\mu = 100$ & 67.9 & {71.3} &41.59\%&17.11\%& 414&\\
$\mu = 1000$& {\bf 71.8} & {\bf 76.3}&{\bf 49.55\%} &{\bf 13.27}\% & 309\\
$\mu = 2000$ & 67.7 & {72.7} & 41.59\% & 16.52\%&428\\
$\mu = 4000$ &{67.5} & 70.5 &41.30\% & 16.81\% &{446}\\
\bottomrule
\end{tabular}
\end{center}
\end{table}

\subsection{Effect of the mutual constraints}
The mutual constraints are used to build connections between crowd density map and object detection. Crowd density map usually has a strong counting ability in crowd scenes, and it thus can provide an informative clue for object detection. In return, object detection can also improve the localization ability of crowd density map. We thus evaluate the effects of ${{\mathcal{L}}_{\text{dc}}}$ and ${{\mathcal{L}}_{\text{cd}}}$ on the validation data set. Also, we validate the effectiveness of the counting task, i.e., ${{\mathcal{L}}_{\text{cnt}}}$. In Tab.~\ref{table:mutual}, we report the tracking results of different variants. When the ${{\mathcal{L}}_{\text{cnt}}}$ is removed,  the CountingMOT model will degrade into FairMOT which doesn't borrow prior information from counting. Thus, FairMOT achieves relatively worse tracking results. When FairMOT is only added with the counting task ${{\mathcal{L}}_{\text{cnt}}}$ (without ${{\mathcal{L}}_{\text{dc}}}$ and ${{\mathcal{L}}_{\text{cd}}}$), MOTA and IDF1 can be improved (e.g., MOTA of 69.1 vs 69.9), which indicates that the counting task implicitly improves the tracking performance. When only  ${{\mathcal{L}}_{\text{dc}}}$ is removed from CountingMOT (denoted as ``w/o ${{\mathcal{L}}_{\text{dc}}}$"), the model can also slightly improve MOTA. In contrast, when only ${{\mathcal{L}}_{\text{cd}}}$ is removed, the model can achieve the highest MT and lowest ML, which benefits from the crowd density map to find missed detections. Overall, the mutual constraints can achieve the best results, making the CountingMOT robust to crowded scenes.

\CUT{
\begin{table}
\begin{center}
\setlength{\tabcolsep}{5pt}
\caption{Effect of mutual constraints ${{\mathcal{L}}_{\text{dc}}}$ and ${{\mathcal{L}}_{\text{cd}}}$ on the validation set of \emph{MOT17}. The best results are shown in {\bf bold}.}
\label{table:mutual}
\begin{tabular}{lcccccc}
\toprule
Mutual constraints & MOTA$\uparrow$ & IDF1$\uparrow$ & MT$\uparrow$& ML$\downarrow$ & IDS $\downarrow$\\
\midrule
w/o ${{\mathcal{L}}_{\text{cnt}}}$ & 69.1 & 72.8 &42.18\%& 15.63\% & {\bf 299}\\
w/o ${{\mathcal{L}}_{\text{dc}}}$, ${{\mathcal{L}}_{\text{cd}}}$, & 69.9 & 72.9&43.07\%& 16.22\% & 325\\

w/o ${{\mathcal{L}}_{\text{dc}}}$ & 70.6 & 72.6 &48.67\%& 14.74\% & 410\\
w/o ${{\mathcal{L}}_{\text{cd}}}$ & 71.1 & {73.8} &{\bf 51.92\%}&{\bf 12.98\%}& 468&\\
${{\mathcal{L}}_{\text{dc}}}$ + ${{\mathcal{L}}_{\text{cd}}}$ & {\bf 71.8} & {\bf 76.3}&{49.55\%} &{13.27}\% & 309\\
\bottomrule
w/o ${{\mathcal{L}}_{\text{id}}}$ (two-stage) & 71.0 & 73.4 &43.07\%& 15.04\% & {389}\\
\bottomrule
\end{tabular}
\end{center}
\vspace{-0.3in}
\end{table}
}

\begin{table}
\begin{center}
\setlength{\tabcolsep}{5pt}
\caption{Effect of mutual constraints ${{\mathcal{L}}_{\text{dc}}}$ and ${{\mathcal{L}}_{\text{cd}}}$ on the validation set of \emph{MOT17}. The best results are shown in {\bf bold}.}
\label{table:mutual}
\begin{tabular}{lcccccc}
\toprule
Mutual constraints & MOTA$\uparrow$ & IDF1$\uparrow$ & MT$\uparrow$& ML$\downarrow$ & IDS $\downarrow$\\
\midrule
w/o ${{\mathcal{L}}_{\text{cnt}}}$ & 69.1 & 72.8 &42.18\%& 15.63\% & {\bf 299}\\
w/o ${{\mathcal{L}}_{\text{dc}}}$, ${{\mathcal{L}}_{\text{cd}}}$, & 69.9 & 72.9&43.07\%& 16.22\% & 325\\

w/o ${{\mathcal{L}}_{\text{dc}}}$ & 70.6 & 72.6 &48.67\%& 14.74\% & 410\\
w/o ${{\mathcal{L}}_{\text{cd}}}$ & 71.1 & {73.8} &{\bf 51.92\%}&{\bf 12.98\%}& 468&\\
${{\mathcal{L}}_{\text{dc}}}$ + ${{\mathcal{L}}_{\text{cd}}}$ & {\bf 71.8} & {\bf 76.3}&{49.55\%} &{13.27}\% & 309\\
\bottomrule
w/o ${{\mathcal{L}}_{\text{id}}}$ (two-stage) & 71.0 & 73.4 &43.07\%& 15.04\% & {389}\\
\new{w/o ${{\mathcal{L}}_{\text{dc}}}$, ${{\mathcal{L}}_{\text{cd}}}$ + TBC} & \new{70.0} & \new{74.3} &\new{42.18\%}& \new{15.93\%} & \new{334}\\
\bottomrule
\end{tabular}
\end{center}
\vspace{-0.3in}
\end{table}

To evaluate the effect of reID 
\new{branch}, we first train a variant of CountingMOT with only detection and counting tasks (denoted as ``w/o ${{\mathcal{L}}_{\text{id}}}$"). Then, we use the ROI-Align strategy~\cite{voigtlaender2019mots} to extract the feature of each bounding box from the backbone features. Finally, we use a fully connected layer to classify these bounding boxes and obtain reID features. This experiment is also conducted on the validation set of MOT17, and the result is reported in Tab.~\ref{table:mutual} (the last row). As observed, the two-stage model ``w/o ${{\mathcal{L}}_{\text{id}}}$" performs worse than our original CountingMOT (IDF1 of 73.4 vs 76.3), which indicates that the joint optimization of reID task can improve the tracking performance. 

\new{TBC [10] also adopts crowd density map to improve object detections,
but it splits density map estimation and object detection
into two separate steps. This means that object detections highly depend on the quality of crowd density map, and they
can’t be optimized simultaneously, which affects the tracking
performance. Here, we also compare TBC with CountingMOT on the validation set of \emph{MOT17}. For a fair comparison, we use CountingMOT without the mutual constraints ${{\mathcal{L}}_{\text{dc}}}$, ${{\mathcal{L}}_{\text{cd}}}$ to generate the input (detections, density maps and reID features) for TBC. As observed in Tab.~\ref{table:mutual}, from MOTA and IDF1, TBC works better than CountingMOT without ${{\mathcal{L}}_{\text{dc}}}$ and ${{\mathcal{L}}_{\text{cd}}}$, and this is because TBC can use information across the whole video to perform data association. When ${{\mathcal{L}}_{\text{dc}}}$ and ${{\mathcal{L}}_{\text{cd}}}$ are added, CountingMOT can optimize object detection and crowd density map simultaneously, and thus achieves better results compared with TBC (e.g., MOTA of 71.8 vs 70.0).}

\begin{figure}[htbp]
\centering
\centering
\includegraphics[width={0.37\textwidth}]{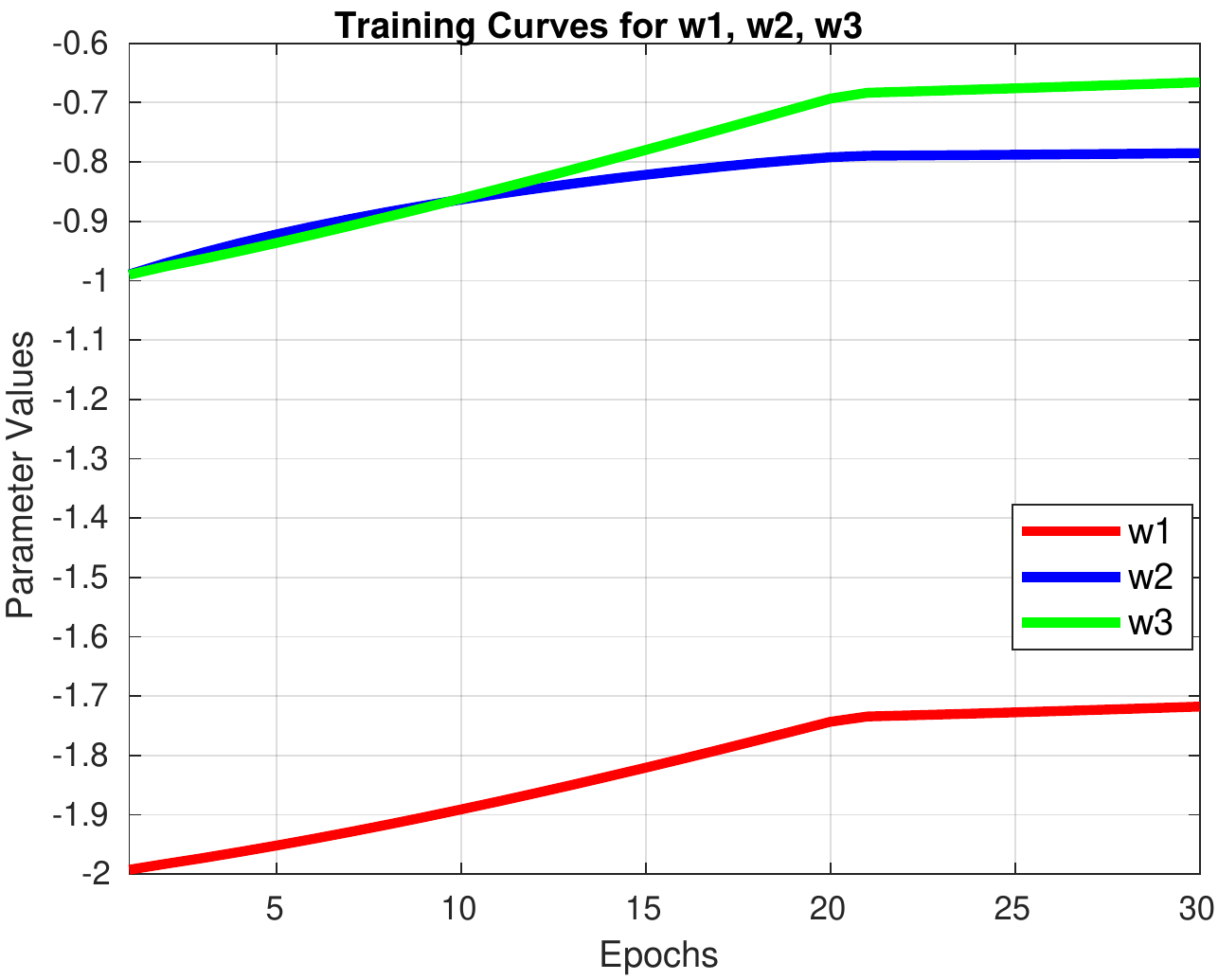} 
\caption{Training Curves for $w_1$, $w_2$, $w_3$.}\label{fig:para}
\vspace{-0.1in}
\end{figure}

\begin{figure}[htbp]
\centering
\centering
\includegraphics[width={0.37\textwidth}]{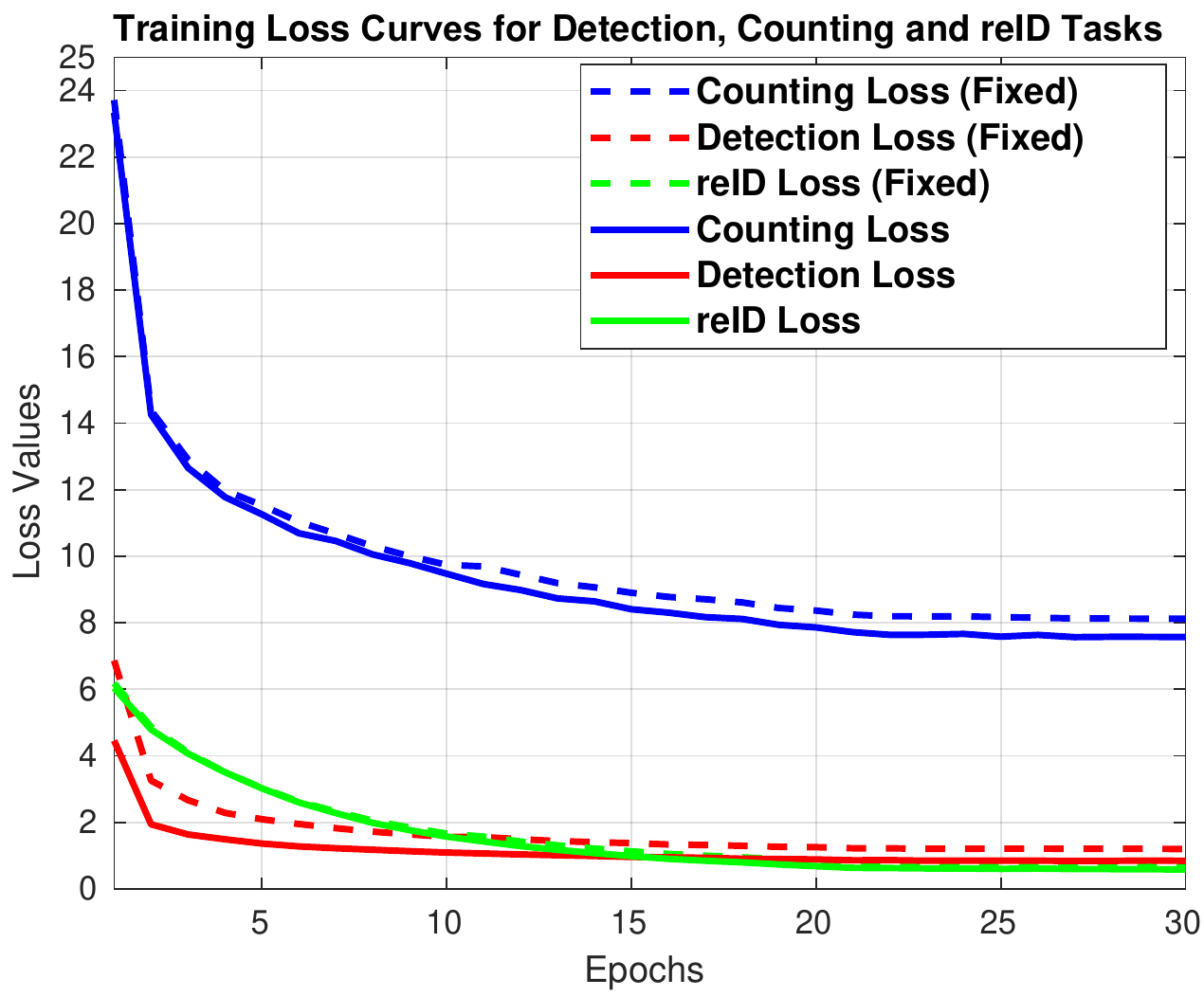} 
\vspace{-0.1in}
\caption{Training \new{loss curves} for detection, counting an reID tasks.}\label{fig:loss}
\vspace{-0.25in}
\end{figure}

\subsection{Effect of the training parameters}
In (\ref{eq:all}), $w_1$, $w_2$ and $w_3$ are trainable parameters to automatically balance the detection, counting and reID tasks. For initialization, we set $w_1$, $w_2$ and $w_3$ to $-2$, $-1$ and $-1$, respectively. In Fig.\ref{fig:para}, we show the training curves for the three parameters on the training set, and they coverage at the 20th epoch. In Fig.~\ref{fig:loss}, the corresponding detection, counting and reID losses also converge at the same epoch. Meanwhile, we show the loss curves for the three tasks (dotted lines in Fig.~\ref{fig:loss}) when the training parameters are fixed ($w_1=-2$, $w_2=-1$ and $w_3=-1$). As observed, the dotted lines converge at relatively high loss values, which indicates that the uncertainty weights are effective for multi-task training. On the validation set, automatically adjusting the three parameters also performs better than using fixed training parameters (MOTA of 71.8 vs 70.2, IDF1 of 76.3 vs 72.3).

The parameter $w_1$ controls the weight of object detection task, and its initial value may affect the tracking performance. To analyze the effect of $w_1$, we initialize $w_2 = -1$, $w_3 = -1$, and change $w_1$ from $-5$ to $-1$ for different training models. The results of different parameter settings are reported in Tab.~\ref{table:para_w}, and we find that a smaller $w_1$ (larger weight for detection) can't improve the tracking performance. The reason is that the detection task is also constrained by the counting task. For the reID weight $w_3$, according to our experiments, the tracking performance has little changes when it varies from $-5$ to $-1$. 

\begin{table}[!htb]
\begin{center}
\setlength{\tabcolsep}{4pt}
\caption{Tracking results of different $w_1$ on the validation set of \emph{MOT17}. }
\label{table:para_w}
\begin{tabular}{lcccccc}
\toprule
 $w_2 = -1$, $w_3 = -1$ & MOTA$\uparrow$ & IDF1$\uparrow$ & MT$\uparrow$& ML$\downarrow$ & IDS $\downarrow$\\
\midrule
$w_1 = -5$ & {60.2} & { 69.3}&{41.89\%} &{16.22}\% & 482\\
$w_1 = -4$ & 68.4 & {71.9} &{43.66\%}&{16.22\%}& 476&\\
$w_1 = -3$ & 71.3 & 74.1 &48.67\%& 13.57\% & 374\\
$w_1 = -2$ & {\bf 71.8} & {\bf 76.3}&{49.55\%} &{\bf 13.27}\% & {\bf 309}\\

$w_1 = -1$ & 70.7 & 72.7 &{\bf 51.33\%}& 15.93\% & {391}\\
\bottomrule
\end{tabular}
\end{center}
\vspace{-0.35in}
\end{table}

\subsection{Evaluation of crowd counting performance}
\new{
To analyze the counting performance, we evaluate different models on the validation sets of \emph{MOT17} and \emph{MOT20} using MAE~(Mean Absolute Error) and SSIM~(Structure Similarity Index Measure). 
The validation sets are generated by uniformly splitting the original train sets (see \ref{sec:IV-D}). The counting results are reported in Tab.~\ref{table:counting}. ``CountingModel" shares the same backbone with FairMOT, but it only has a counting branch. ``FairMOT+Counting" is created by adding the additional counting task to FairMOT. Note that all the models in Tab.~\ref{table:counting} are first pre-trained on the extra datasets (see \ref{sec::datasts}), and then are fine-tuned on the validation sets. As observed in Tab.~\ref{table:counting}, for \emph{MOT17}, the CountingMOT achieves the highest MAE and SSIM. For the crowd scene \emph{MOT20}, the CountingMOT has the highest SSIM, which implies that the estimated density map has a better localization ability. In summary, our CountingMOT model can improve FairMOT from the counting results, and it can also improve the localization ability of crowd density map in crowd scenes (from SSIM).}

\begin{table}
\begin{center}
\setlength{\tabcolsep}{5pt}
\caption{Crowd counting performance of different models on the validation sets of \emph{MOT17} and \emph{MOT20}.}
\label{table:counting}
\begin{tabular}{|l|cc|cc|}
\hline
\multirow{2}{*}{Models} &\multicolumn{2}{c|}{MOT17}&\multicolumn{2}{c|}{MOT20} \\
\cline{2-5}
&MAE&SSIM&MAE&SSIM \\
\hline
FairMOT&20.56&-&17.08&- \\
CountingModel&18.04&{0.84}&\textbf{16.11}&0.70 \\
FairMOT+Counting&18.36&0.82&16.80&0.68 \\
CountingMOT&\textbf{17.55}&\textbf{0.87}&16.27& \textbf{0.72} \\
\hline
\end{tabular}
\end{center}
\vspace{-0.3in}
\end{table}

\section{Conclusion}
\label{text:conclusion}
In this paper, we propose a multi-task model to jointly perform crowd counting, detection and re-Identification. Unlike the existing \emph{tracking-by-detection} MOT methods, our CountingMOT model introduces the informative crowd density map to help recover missed or occluded object detections, which makes our model robust to crowd scenes. Also, the detection task can improve the quality of crowd density map. The mutual constraints between detection and counting can jointly improve the tracking performance. Our approach is an attempt to bridge the gap of counting, detection and re-Identification. Experimental results show that our model achieves the state-of-the-art results on \emph{MOT16} and \emph{MOT17}. Also, the comparison on \emph{MOT20} indicates that our CountingMOT model can well solve multi-object tracking in crowd scenes.
\new{In the future, we will further improve the ReID task with unsupervised or semi-supervised learning~\cite{zhou2019person,chen2019semisupervised,gu2022motion,zhou2021multinetwork}. For real applications, we will also consider the long-term association by jointly using the correlation filtering~\cite{hu2017manifold,yuan2020self,lukezivc2020performance}}




\ifCLASSOPTIONcaptionsoff
  \newpage
\fi

\footnotesize
\bibliographystyle{IEEEtran}

\bibliography{IEEEexample}

\vfill

\end{document}